\documentclass[runningheads]{llncs}

\usepackage{eccv}

\usepackage{eccvabbrv}

\usepackage{graphicx}
\usepackage{booktabs}
\usepackage{adjustbox}

\usepackage[accsupp]{axessibility}  % Improves PDF readability for those with disabilities.

\usepackage{hyperref}

\usepackage{orcidlink}
\usepackage{algorithmic}
\usepackage{algorithm}
\usepackage{float}
\usepackage{amsmath}
\usepackage{booktabs}
\usepackage{multirow}
\usepackage{graphicx}

\usepackage{subcaption}
\usepackage{caption}
\usepackage{array}
\usepackage{placeins}

\begin{document}

% \title{AnyBox: Efficient Zero-Shot Category-Level 6D Pose Estimation of Boxes for Warehouse Robotics} 

% \title{AnyBox: Efficient Zero-Shot Category-Level 6D \textit{Any} Box Pose Estimation} 
\title{AnyBox: Efficient Zero-Shot 9DoF Pose Estimation of Boxes for Robotic Manipulation \vspace{-0.5cm}}

\titlerunning{AnyBox: Efficient Zero-Shot 9DoF Pose Estimation of Boxes}

% TODO FINAL: Include the authors' ORCID for the camera-ready version, if at all possible.
\author{Yintao Ma\textsuperscript{1,*} \and
Sajjad Pakdamansavoji\textsuperscript{1,*,\dag,\ddag} \and
Charles Eret\textsuperscript{1,2,\ddag} \and
Rui Heng Yang\textsuperscript{1} \and
Xuan Zhao\textsuperscript{1} \and
Yingxue Zhang\textsuperscript{1} \and
Tongtong Cao\textsuperscript{1} \and
Amir Rasouli\textsuperscript{1}}

\authorrunning{Y.~Ma, S.~Pakdamansavoji, et al.}

\institute{\textsuperscript{1}Huawei Technologies Canada \qquad
\textsuperscript{2}University of Waterloo\\[2pt]
\textsuperscript{*}Equal contribution \quad
\textsuperscript{\dag}Corresponding author \quad
\textsuperscript{\ddag}Work done while at Huawei}

\maketitle

\vspace{-0.5cm}
\begin{abstract}
Recovering the 9D pose of objects, both their 6D pose and 3D dimensions, under clutter and occlusion is a core requirement for warehouse automation, logistics, and manufacturing. Model-based methods are accurate but assume an instance-specific CAD model for every object, which is costly to maintain as inventories change. Model-free and category-level methods relax this assumption, yet they remain vulnerable to the symmetry, weak texture, and heavy occlusion that characterize stacked storage boxes, and they ignore the strong structural priors such scenes provide. We present \textbf{AnyBox}, an efficient zero-shot framework that exploits the geometric regularity of boxes to jointly recover pose and dimensions from a single RGB-D observation. Starting from a canonical category template, AnyBox alternates between pose and scale estimation, using the discrepancy between the reprojected template and the observed mask to drive a binary search over box dimensions. Two lightweight components make this practical: a depth-consistency filter that rejects the implausible hypotheses induced by box symmetry, and an early-stopping rule that replaces the remaining search with a single closed-form update. On public benchmarks and an in-house warehouse dataset, AnyBox improves detection AP by up to 36 points, more than doubling the previous best, and approaches instance-level pipelines that have access to ground-truth CAD models. These gains transfer downstream, raising success by 28\% on a cluttered robotic box-shelving task.

\end{abstract}

\section{Introduction}
\label{sec:intro}

% \begin{figure*}[t]
%     \centering
%     % \includegraphics[width=\textwidth]{figures/overview.png}
%     \includegraphics[width=\textwidth]{figures/images/Schematic_Overview.jpeg}
%     \caption{Schematic overview of zero-shot category-level 9D pose estimation of boxes with AnyBox in an industrial warehouse environment.}
%     \label{fig:sch_overview}\vspace{-0.5cm}
% \end{figure*}

Modern warehouses and distribution centers are increasingly being automated via smart logistics and deployment of robotic systems.  In these settings, large volumes of packaged goods should be handled reliably in highly dynamic and cluttered operating spaces. To function effectively, objects' poses, including  both the 3D position and the 3D orientation, must be accurately and efficiently estimated. This capability is essential for executing fundamental automated tasks, such as robotic manipulation of boxes involving picking, placing, sorting, and transporting. In such cases, even small errors in position or orientation estimation can lead to unstable grasping, collision, or failed placements.

In industrial settings, estimating the pose of boxes is challenging because they often vary in size, appearance, and condition. These variations can introduce visual distractions or cause partial occlusions. In addition, a robust system must handle previously unseen objects without relying on exact models.

The existing 6D pose estimation approaches are predominantly model-based.  They assume a high-resolution instance-level CAD model for each target and rely on template rendering at inference. They also require vast instance-level CAD libraries, which could be costly to maintain. This requirement limits the scalability and transferability of these models to new environments. Model-free and category-level methods remedy this problem by using only a few reference frames and generic CAD templates per category, respectively. These approaches, however, are not suitable for warehouse settings where objects are packed in boxes of various dimensions and appearances. Model-free models struggle to accurately estimate poses of boxes due to their symmetric geometry and lack of distinctive visual features \cite{housecat}. Category-level models face another limitation: they do not exploit strong contextual priors, such as common object types or structured layouts. As a result, they are more vulnerable to clutter and occlusion.

To this end, we propose \textbf{AnyBox}, a zero-shot category-level 9D pose estimation pipeline tailored for boxes in industrial settings. Specifically, we extend upon model-based approaches that rely on instance-specific CAD models. Instead of assuming access to the exact geometry of each object, our approach uses a novel scale estimation module that infers the size of boxes directly from a single RGB-D observation using canonical templates. Our model operates in an iterative two-stage format. It starts with a pose estimation pass, which identifies the best pose candidate, followed by scale estimation to evaluate the observed mask against the best candidate's reprojection. If the matching condition is not satisfied, the discrepancy between the candidate and the observed masks is used to scale the templates to be fed back into the pose estimation module. The iterative estimation and scaling process is implemented as a binary search over possible dimensions, enabling efficient convergence to an accurate scale and pose.

We address geometric ambiguity challenges of estimating box-shaped objects' poses using a filtering process. Since multiple pose hypotheses can appear visually plausible under partial observations, a depth-based plausibility filter is applied to evaluate geometric consistency between the reprojected hypotheses and the observed depth map. Hypotheses that violate depth constraints are discarded early, reducing false positives and improving stability under occlusion and clutter. Finally, we integrate an early-stopping strategy into the search procedure, terminating the refinement process once improvements become marginal. This design reduces computational cost while preserving pose accuracy, resulting in a fast and reliable system suitable for real-world tasks. \noindent In summary, the contributions of this work are:
\begin{itemize}
  \vspace*{-0.2cm}
    % \item We propose a novel 9D pose estimation framework for boxes in industrial settings. Our method uses an iterative two-stage process that estimates and scales pose templates until the best match is resulted. Our method incorporates a novel depth-based plausibility filter that enforces geometric consistency to resolve ambiguities arising from object symmetry, partial observations and clutter. An early-stopping mechanism further minimizes the computational overhead of the search process. 
    \item We propose \textbf{AnyBox}, a novel zero-shot category-level framework for joint 9D pose estimation of box-shaped objects. Unlike existing category-level methods, AnyBox generalizes across diverse warehouse environments without environment-specific training or fine-tuning through iterative joint pose and dimension refinement from a single RGB-D observation.

    \item We conduct extensive evaluations on both our in-house warehouse dataset and common pose estimation benchmarks to validate the effectiveness of our approaches compared to state-of-the-art methods.

    \item We evaluate the robustness of our approach in a downstream box-shelving robotic manipulation task in highly cluttered environments. 

    \item We provide comprehensive ablation studies to quantify the individual contributions of the proposed scale estimation, depth plausibility filtering, and early-stopping modules to the overall system performance.

\end{itemize}

\section{Related Work}
In automated logistics settings, a wide variety of objects should be perceived and manipulated under challenging conditions involving clutter, partial occlusion, and strict runtime constraints. Accurate recovery of both 3D position and 3D orientation is essential for reliable grasp planning and execution. However, estimating the pose of unseen objects, especially those with strong geometric symmetry and weak texture, remains challenging. Existing approaches for object pose estimation can be broadly organized based on their use of object CAD models at inference time.
\vspace{-0.5cm}
\subsubsection{Instance-Level Model-Based Methods.}
Traditional 6D pose estimation methods assume that the exact CAD model of each object is available at test time. These approaches tightly couple image observations with known geometry, often achieving high accuracy. Early systems are trained per object instance to learn stable 2D--3D cues, such as keypoints~\cite{peng2019pvnet}, view or rotation labels~\cite{xiang2018posecnn}, dense surface representations~\cite{haugaard2022surfemb}, or geometry-guided predictors~\cite{wang2021gdr,su2022zebrapose,yang2023scflow}. Alternatively, some models rely on render-and-compare strategies~\cite{coordicate} or point cloud alignment with SVD-based solvers~\cite{megapose}. When the object is seen during training and its CAD model is known, these methods can achieve strong performance even under heavy occlusion because correspondence supervision anchors image evidence directly to the specific mesh~\cite{hodan2020epos}. Instance-level model-based methods, however, face scalability limitations in warehouse applications. Maintaining high-quality CAD models and template banks for every box instance is costly, if not infeasible, especially when new products are introduced frequently. Moreover, transferring such systems across warehouses or updating object inventories requires additional modeling and calibration effort, limiting their practical deployment at scale.
\vspace{-0.5cm}
\subsubsection{Model-Free and Few-Shot Methods.}
To improve generalization to unseen objects, model-free approaches remove the requirement for CAD models at inference. Some methods learn mappings from images directly to pose or to intermediate object-centered representations~\cite{pixel2pose}. Others construct an implicit reference model from a small set of images or short videos and estimate pose through feature matching and refinement~\cite{sun2022onepose,liu2022gen6d,any6d,fp}.  While these strategies offer flexibility, they struggle in warehouse environments. Box-shaped objects often exhibit strong symmetry, repetitive appearance, and limited texture. Furthermore, under clutter and occlusion conditions, local feature matching becomes unreliable, and sparse references provide insufficient geometric constraints. As a result, accuracy degrades, particularly when real-time processing is required.
\subsubsection{Category-Level Methods.}
Category-level 6D pose estimation aims to bridge the gap between instance-level accuracy and model-free flexibility. Instead of using an instance-specific mesh, these methods rely on a shared canonical representation for all objects within a category. A representative example is NOCS~\cite{Wang_2019_CVPR}, which predicts normalized object coordinates and recovers pose and scale through alignment. Subsequent works extend this framework by learning to deform a category-level shape prior to better match each instance~\cite{tian2020shape} or by improving correspondence prediction and shape reconstruction~\cite{chen2021sgpa,lin2022category, wang2021category, zhang2022rbp}.

Although category-level approaches improve generalization, several challenges remain for industrial deployment. First, symmetric objects introduce pose ambiguity that generic canonical representations do not explicitly resolve. Second, intra-category scale variation can lead to inaccurate alignment when size is not explicitly modeled. Third, many existing methods are designed for broad object categories and do not exploit strong structural priors present in warehouse settings. For example, storage boxes follow simple geometric rules, have constrained aspect ratios, and are often arranged in structured layouts. Ignoring these regularities limits robustness under symmetry, occlusion, and weak appearance cues.

\begin{figure*}[t]
    \centering
    \includegraphics[width=\textwidth]{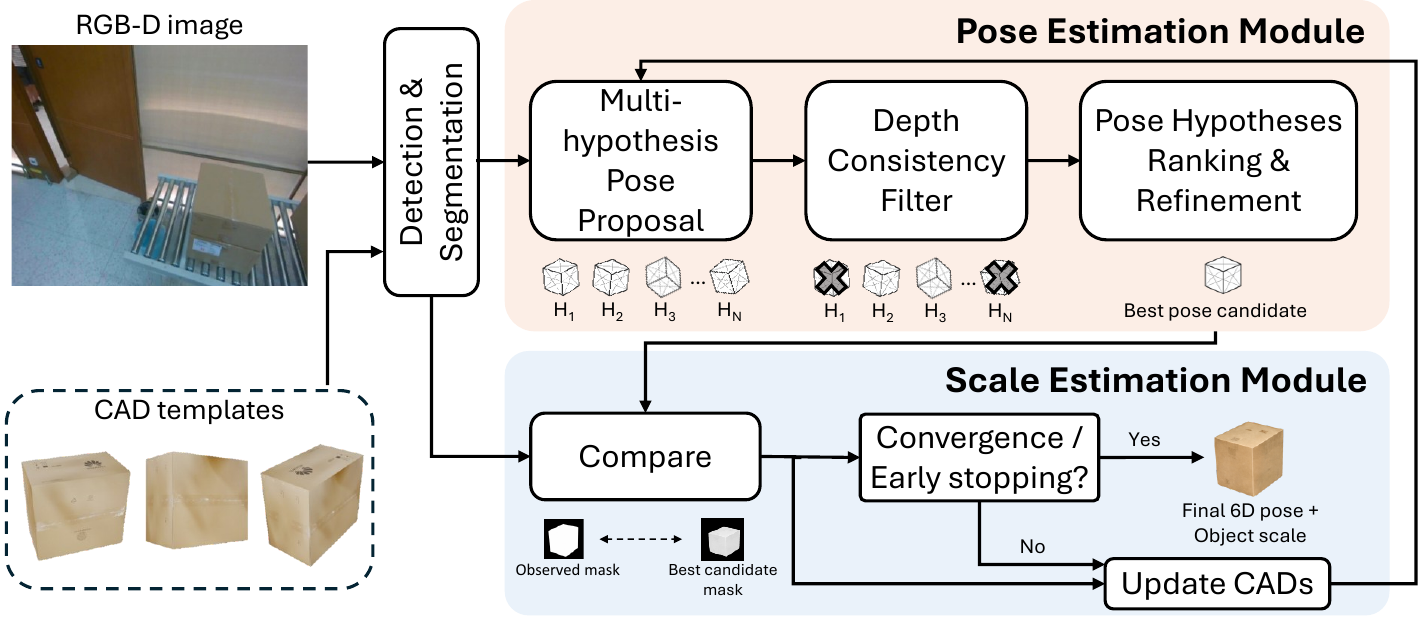}
    \caption{Overview of the AnyBox pipeline. The observed RGB–D image and a rendered RGB–D category box template are first used for detection and segmentation. The detections are passed to the pose estimator to generate multiple pose hypotheses, reject those with depth inconsistency, and refine the high-confidence candidates. The scale estimation module uses the resulting pose to guide dimension estimation by comparing the reprojected CAD mask to the observation. Then, based on their discrepancy, the CAD templates are rescaled accordingly and fed back to the pose estimation module. This project–compare–rescale loop is implemented as a binary search, iteratively refining the scale estimates by narrowing the search interval until convergence or termination via early-stopping.}
    \label{fig:overview}
    \vspace{-0.5cm}
\end{figure*}

\section{Method}
In this work, we focus on category-level 6D pose and 3D dimension estimation tailored to box-shaped objects. Our goal is to combine the geometric consistency of model-based methods with the flexibility of category-level formulations, while explicitly addressing scale variation and symmetry challenges. By leveraging the structural simplicity of boxes and incorporating domain-specific constraints, we aim to develop a pose estimation framework that is accurate, efficient, and practical for large-scale robotic logistics systems.

Our proposed framework, AnyBox, consists of three components: \textbf{Object detection}, which localizes the target and produces a segmentation mask; \textbf{Pose estimation}, which generates multiple pose hypotheses via point-cloud matching, prunes them via a \textit{depth-consistency filter}, and ranks and selects the best candidate; \textbf{Scale estimation}, which iteratively estimates the box dimensions along the X, Y, and Z axes, rescales the CAD templates, and repeats this process until convergence or termination via \textit{early stopping}. An overview of the framework is shown in Fig. \ref{fig:overview}. Details of each module are provided below.

\subsection{Object Detection}
Following \cite{lin2024sam6d}, we prompt Segment Anything Model (SAM) \cite{kirillov2023sam} to produce class-agnostic mask proposals on the observed image, filter them by confidence and non-maximum suppression (NMS), and then apply an object-matching score to only keep the proposals that are consistent with the target instance \cite{kirillov2023sam,lin2024sam6d}. The matching combines high-level semantics with coarse geometric consistency, without requiring per-class training \cite{lin2024sam6d,oquab2023dinov2}. The retained proposals provide the bounding boxes and segmentation masks  passed to the pose estimation stage \cite{lin2024sam6d}.

\subsection{Pose Estimation}
\subsubsection{Hypothesis Generation.}
We estimate pose via 3D–3D correspondences between the observation and the generic category model point clouds. Following \cite{lin2024sam6d}, a lightweight proposal network produces multiple initial pose hypotheses. We then apply a depth-consistency validity check to prune implausible hypotheses. From the remaining set, we select the highest-confidence hypothesis and feed it to a refinement network that further aligns the point clouds to sharpen rotation and translation estimations.
\vspace*{-0.5cm}
\subsubsection{Depth-Consistency Filter.}
\begin{figure}[t]
  \centering
  \includegraphics[width=1\linewidth]{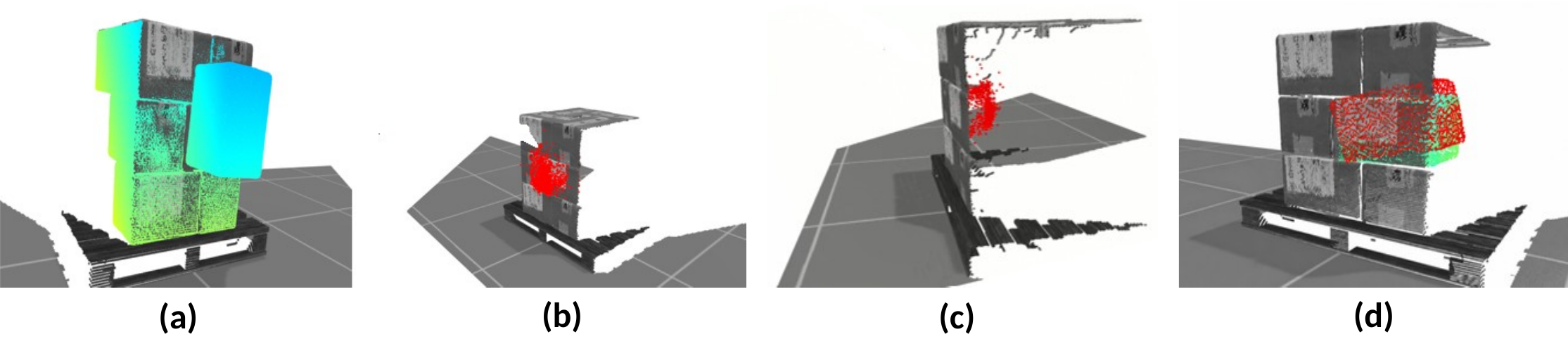}\vspace{-0.2cm}
  \caption{Visualizing the depth-consistency filter for 9D box pose precision. 
  \textbf{(a)} A rejected pose hypothesis showing inconsistent depth protruding from the stack. 
  \textbf{(b)} Centers of all initial pose hypotheses (red), illustrating the distribution of candidate poses before depth-consistency filtering.
  \textbf{(c)} Remaining pose hypotheses centers after depth-consistency filtering, correctly localized within the stack.
  \textbf{(d)} Final estimated 9D pose (red) vs. ground-truth (green) after filtering.}
  \label{fig:vis_depth}
  \vspace{-0.5cm} % Standard tweak for top-tier conference spacing
\end{figure}

Boxes are often symmetric along several axes and their textures are weak or repetitive, so hypotheses with incorrect rotations can obtain confidence scores similar to correct ones in the pose-proposal stage. A wrong rotation typically swaps the visible face with an opposite face, which makes the reprojected geometry inconsistent with the observed depth; this ambiguity is common for stacked or occluded boxes, where coarse poses can appear to “hover” above the true surface (see Fig.~\ref{fig:vis_depth}-(a)). We address this with a novel depth-consistency filter: for each hypothesis, we reproject the CAD model into the camera frame and compare its depth with the observed depth inside the predicted mask. Specifically, we partition the observed depth into $5\times5$ pixel cells, record the minimum depth per cell, and project $1{,}000$ uniformly sampled CAD points into the image. Points whose projected depth lies in front of the corresponding cell depth, using a point-wise threshold of zero, contribute to a depth-consistency penalty that re-ranks the hypotheses. The remaining high-confidence hypothesis enables a more reliable refinement stage.

\subsubsection{Refinement and Ranking.}
Once consistent pose hypotheses are selected, a refinement and ranking stage optimizes the initial pose estimates using a larger, higher-capacity network \cite{lin2024sam6d} that is structurally similar to the pose hypothesis generation model. Explicitly conditioning this network on the initial hypothesis lets it focus on extracting fine-grained adjustments. Alongside the refined pose parameters, the network outputs a confidence score, which is used to rank all hypotheses and select the most confident final pose.

\subsection{Scale Estimation}
\begin{figure}[t]
\centering
\includegraphics[width=0.9\linewidth]{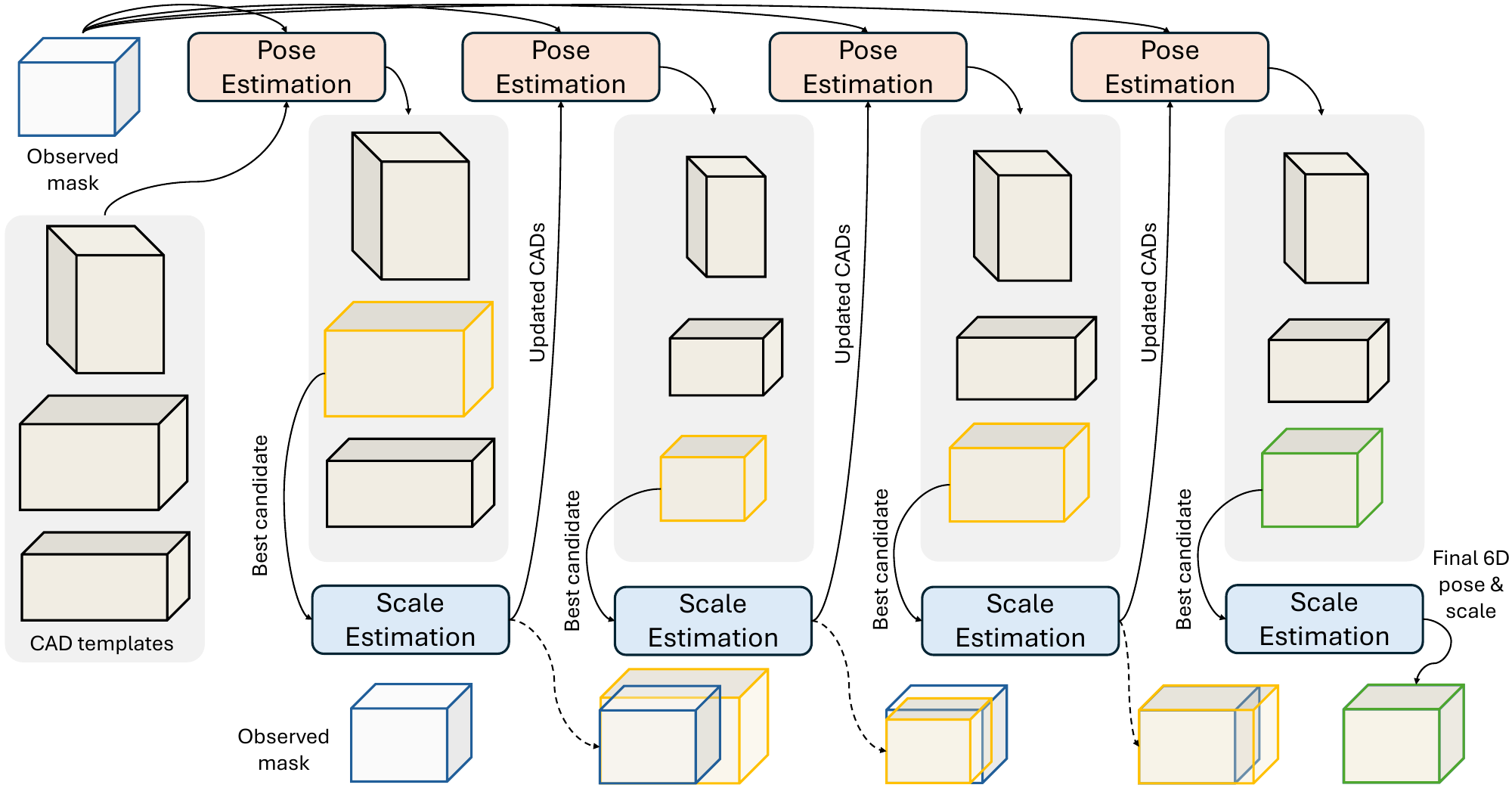}
\caption{Overview of the binary search process. The templates are ranked and the best candidate is selected and passed to the scale estimation module. If it does not converge with the observed mask, templates are scaled and returned to the pose estimation module. This process continues until a candidate converges with the observed mask.}
\label{fig:dimension_estimation}\vspace{-1cm}
\end{figure}

We estimate box dimensions using a canonical, category-level CAD template and the 6D pose recovered by the pose estimation module. With this pose fixed for the current iteration, we scale the CAD independently along the X, Y, and Z axes, place the scaled model at the estimated pose, and reproject its vertices into the camera using the known intrinsics to create a synthetic mask. We then compare the synthetic mask with the observed mask by measuring their axis-aligned pixel extents along the image directions that correspond to the three box axes. For each axis, the comparison gives a true-false decision: if the CAD underfills the observation we increase the scale on that axis, and if it overfills we decrease it (Fig. \ref{fig:dimension_estimation}). We implement this as a binary search over scale on each axis, updating the lower or upper bound based on the decision and setting the next hypothesis to the midpoint of the interval. After each scale update, we refresh the 6D pose using the updated CAD so that coupling between scale and pose does not accumulate and the alignment remains stable. This loop is related to iterative closest point (ICP) in that both alternate between fitting a model and updating it, but ICP refines only the rigid transform of a fixed-size model, whereas our procedure also recovers the box dimensions from the mismatch between the reprojected and observed masks.

The process repeats, reprojects, compares, decides, updates bounds, and re-estimates the pose, until convergence or an early-stopping criterion is met. The stopping criterion can be defined in several ways, such as achieving a small per-axis extent error (e.g., 10 pixels), reaching sufficiently tight search intervals, or hitting a maximum number of iterations; in our implementation, we adopt the first criterion. In all of our experiments the search always terminated, either by meeting this criterion or by reaching the iteration limit, although, as with any iterative procedure, the result is not guaranteed to be optimal and inaccurate intermediate poses can lead to a suboptimal final estimate.
 See Algorithm \ref{alg:binary-search}.

\begin{figure}[t]
\centering
\includegraphics[width=\linewidth]{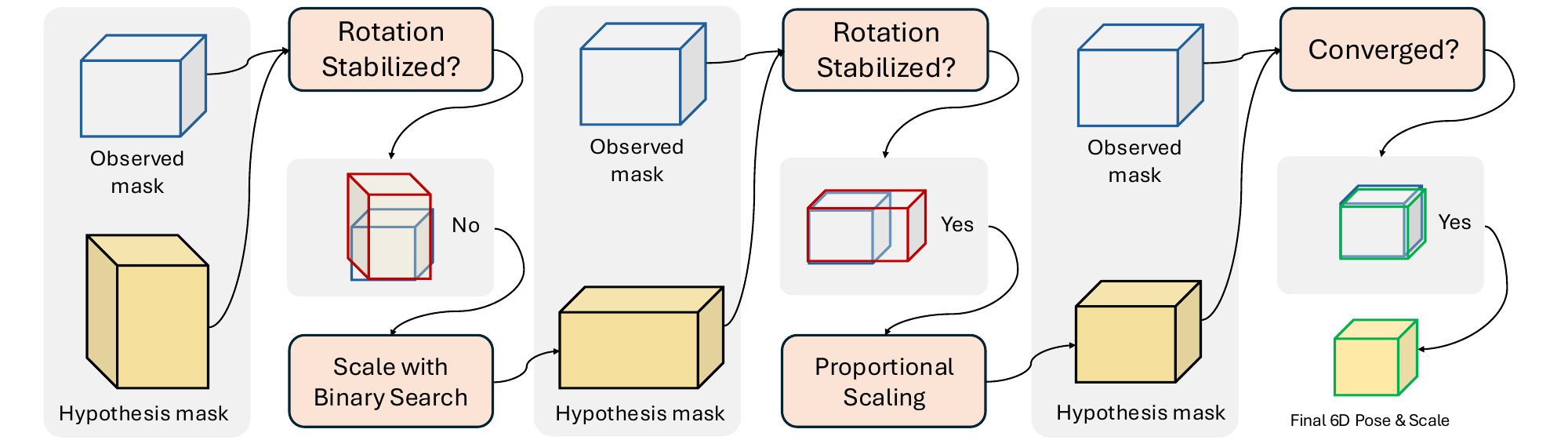}
\caption{Overview of the early stopping mechanism. Once the rotation converges and the reprojected hypothesis mask aligns with the observed mask, the remaining scale ambiguity is resolved using a proportional update based on the ratio of mask extents. This replaces the remaining binary-search iterations and terminates the process.}
\label{fig:early-stopping}
\vspace{-0.5cm}
\end{figure}

\textbf{Early Stopping.} Once the estimated rotation has stabilized and the projected box axes are aligned with the CAD template, the remaining uncertainty lies primarily in the object scale. We detect this scenario by monitoring the change in the estimated rotation across iterations and switch when the relative rotation falls below a small threshold for consecutive steps, indicating that the pose estimator has converged in orientation. At this stage, we replace the iterative binary search with a single proportional update. Specifically, we compute the axis-aligned pixel extents of the observed mask and the reprojected CAD mask, take their per-axis ratios, and rescale the CAD template accordingly in one step. A short search phase is still required at the beginning because the pose estimator is sensitive to scale and requires a roughly correct template size to converge to the correct orientation. After this initialization phase, the proportional update effectively provides a closed-form estimate of the final box dimensions, as illustrated in Fig.~\ref{fig:early-stopping}. In practice, we apply the proportional scaling, re-estimate the pose once to absorb any residual coupling between scale and pose, and terminate the procedure. This strategy removes unnecessary binary-search iterations and substantially reduces computation time. See Algorithm ~\ref{alg:early-stopping}.

\begin{figure}[t]
\centering
\begin{minipage}[t]{0.49\textwidth}
\vspace{0pt}
\begin{algorithm}[H]
\footnotesize
\caption{Binary Search for Box Dimensions}
\label{alg:binary-search}
\begin{algorithmic}[1]
\REQUIRE CAD $\mathcal{C}$, intrinsics $K$, observed mask $\mathcal{M}_{\text{obs}}$, initial pose $T_0$, bounds $\ell,u$, tolerance $\tau_{\text{px}}$, max iters $T_{\max}$
\ENSURE Scales $s$ and pose $T$
\STATE $s \gets (1,1,1)$;\ \ $T \gets T_0$
\FOR{$t=1$ to $T_{\max}$}
  \STATE $T \gets \textsc{PoseEstimate}(\mathcal{C}(s))$
  \STATE $\mathcal{M}_{\text{cad}} \gets \textsc{RenderMask}(\mathcal{C}(s), T, K)$
  \STATE $e_{\text{cad}} \gets \textsc{Extents}(\mathcal{M}_{\text{cad}})$
  \STATE $e_{\text{obs}} \gets \textsc{Extents}(\mathcal{M}_{\text{obs}})$
  \IF{$\max_a |\,e_{\text{cad},a} - e_{\text{obs},a}\,| \le \tau_{\text{px}}$}
    \STATE \textbf{break}
  \ENDIF
  \STATE $m \gets e_{\text{obs}} - e_{\text{cad}}$
  \STATE $\ell_a \gets s_a$ if $m_a>0$; else $u_a \gets s_a$
  \STATE $s \gets (\ell + u)/2$
\ENDFOR
\STATE \textbf{return} $s, T$
\end{algorithmic}
\end{algorithm}
\end{minipage}
\hfill
\begin{minipage}[t]{0.49\textwidth}
\vspace{0pt}
\begin{algorithm}[H]
\footnotesize
\caption{Early stopping via proportional scaling}
\label{alg:early-stopping}
\begin{algorithmic}[1]
\REQUIRE CAD $\mathcal{C}$, intrinsics $K$, observed mask $\mathcal{M}_{\text{obs}}$, poses $T_{\text{prev}}, T$, scales $s$, threshold $\tau_{\text{rot}}$
\ENSURE Final scales $s$, pose $T$
\STATE $\Delta\theta \leftarrow \arccos\!\big((\mathrm{tr}(R_{\text{prev}}^\top R)-1)/2\big)$
\IF{$\Delta\theta \le \tau_{\text{rot}}$}
  \STATE $\mathcal{M}_{\text{cad}} \leftarrow \textsc{RenderMask}(\mathcal{C}(s), T, K)$
  \STATE $e_{\text{cad}} \leftarrow \textsc{Extents}(\mathcal{M}_{\text{cad}})$
  \STATE $e_{\text{obs}} \leftarrow \textsc{Extents}(\mathcal{M}_{\text{obs}})$
  \STATE $r \leftarrow e_{\text{obs}} \oslash e_{\text{cad}}$
  \STATE $s \leftarrow s \odot r$
  \STATE $T \leftarrow \textsc{PoseEstimate}(\mathcal{C}(s))$
  \STATE \textbf{return} $s, T$
\ELSE
  \STATE \textbf{return} \STATE \textsc{BinarySearchRefine} \STATE $(\mathcal{C}, K, \mathcal{M}_{\text{obs}}, T, s)$
\ENDIF
\end{algorithmic}
\end{algorithm}
\end{minipage}
\end{figure}

\section{Experiments}
\subsection{ Pose Estimation}
\subsubsection{Datasets.} We evaluate our method across three datasets encompassing diverse challenges: dense clutter, multiple objects, static and dynamic scenes, tabletop robotic manipulation settings, and objects that are textureless, reflective, symmetric, or varied in size.

\emph{HouseCat6D}\cite{housecat} comprises real and synthetic RGB-D data of household scenes with significant clutter and frequent occlusions. It includes multi-instance arrangements and varied surface properties, including low texture and specular reflections. We evaluate on the \emph{box-like} subset relevant to our category-level setting, consisting of \textit{4 scenes} with \textit{2229 object instances}.

\emph{PACE}\cite{pace} focuses on manipulation-centric scenes, with tabletop layouts and robot interactions that induce both static and dynamic occlusion patterns. The dataset covers challenging geometries and materials, including symmetric and reflective objects. We use the \emph{packaging/box} subset for category-level evaluation, comprising \textit{63 test scenes} and \textit{11638 object instances}.

\emph{Warehouse6D} is a \emph{proprietary} in-house RGB-D collection captured in a warehouse and tabletop setups with diverse lighting, background clutter, and occlusion. This data contains multi-view sequences with multiple box instances per scene and motion from both camera and objects.

\subsubsection{Metrics.}
% Following prior work \cite{nocs}, we evaluate 6D pose estimation using 3D Intersection-over-Union (IoU) between the predicted and ground-truth object volumes. For the public benchmarks, we report IoU at thresholds of 25\% and 50\% which are standard on the official leaderboards. For our Warehouse6D dataset, we adopt a stricter evaluation and report IoU at thresholds of 50\%, 70\%, and 90\% in order to assess the geometric accuracy of the estimated poses against orcale models with ground-truth CAD templates. 
Following prior work \cite{nocs}, we evaluate 6D pose estimation using 3D Intersection-over-Union (IoU) between the predicted and ground-truth object volumes. For the public PACE benchmark, we report detection AP at IoU thresholds of 25\% and 50\%, together with pose AP under three accuracy criteria consisting of rotation $\leq20^\circ$, translation $\leq5$ cm, and their conjunction, as well as ADD-S AP at 0.1$d$ and 0.15$d$. For our Warehouse6D dataset, we adopt a stricter evaluation by reporting detection AP at IoU thresholds of 50\%, 70\%, and 90\%. We additionally report pose AP under the same three accuracy criteria and ADD-S AP at 0.1$d$ and 0.15$d$ to provide a comprehensive evaluation of pose accuracy against oracle models with ground-truth CAD templates.

\subsubsection{Models.}
In our evaluation, we follow the official leaderboards of each public benchmark and compare against the state-of-the-art methods reported for those datasets. Specifically, on HouseCat6D, we report comparisons to NOCS \cite{nocs}, FS-Net \cite{fs-net}, GPV-Pose \cite{gpv-pose}, VI-Net \cite{vi-net}, AG-Pose \cite{lin2024instance}, and SecondPose \cite{secondpose}. On PACE, we compare with NOCS \cite{nocs}, HS-Pose \cite{hs-pose}, SGPA \cite{chen2021sgpa}, SAR-Net \cite{sarnet}, and CPPF++ \cite{cppf}, following the respective evaluation protocols and leaderboards. On our Warehouse6D, we use SAM6D as the baseline and perform ablation studies to systematically isolate the contributions of each component in AnyBox.

\vspace{-0.25cm}
\subsubsection{HouseCat6D.}
\begin{table}[t]
\centering
\caption{Comparison of AnyBox with state-of-the-art models on HouseCat6D. 6D pose performance is reported as average precision at IoU = 0.25 and 0.50.}
\label{t:HouseCat6D} 
\begin{tabular}{l|cc}
\toprule
Model   & IoU=.25 \qquad   & IoU=.50 \\
\midrule
FS-Net\cite{fs-net} & 31.7 & 1.2 \\
GPV-Pose\cite{gpv-pose} & 31.4 & 1.1 \\
NOCS\cite{nocs} & 43.3 & 6.5 \\
VI-Net\cite{vi-net} & 44.8 & 12.7 \\
SecondPose\cite{secondpose} & 54.5 & 23.7 \\
AG-Pose\cite{lin2024instance} & 57.2 & 7.7 \\
\textbf{AnyBox (Ours)} \qquad & \textbf{88.8} & \textbf{58.9}\\
\bottomrule
\end{tabular}\vspace{-0.3cm}
\end{table}

\begin{figure}[t]
\centering
  \includegraphics[width=1\linewidth]{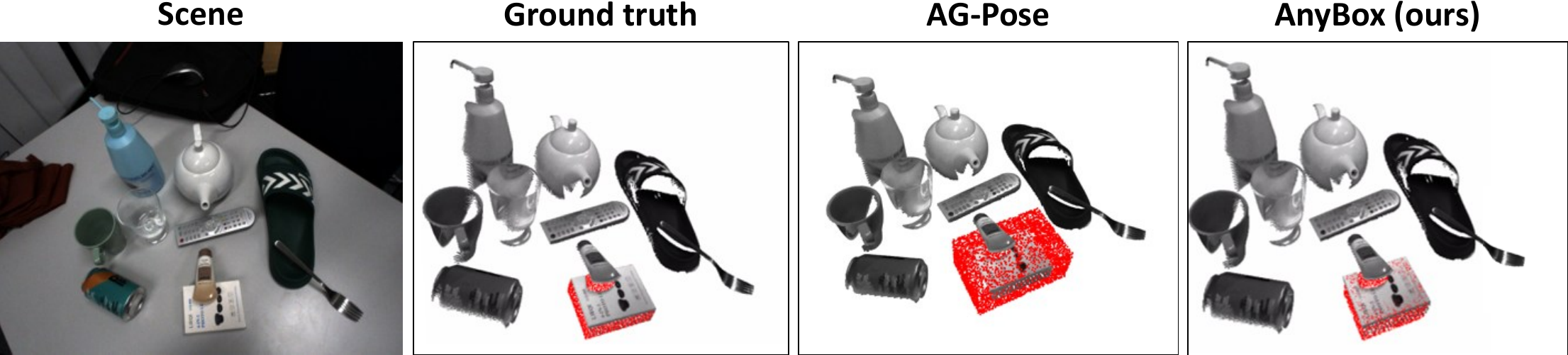}
  \caption{Qualitative evaluation of our model AnyBox, on HouseCat6D dataset \cite{housecat}.} 
\label{fig:example_housecat6D}\vspace{-0.3cm}
\end{figure}

As shown in Table~\ref{t:HouseCat6D}, AnyBox achieves the highest performance at both thresholds, reaching 88.8 at IoU = 0.25 and 58.9 at IoU = 0.50, substantially outperforming the next best method, AG-Pose, which reports 57.2 and 7.7, respectively, highlighting the effectiveness of our approach in accurately recovering object pose and geometry for unseen box instances. Qualitative examples in Fig.~\ref{fig:example_housecat6D} illustrate that while AG-Pose often recovers accurate translation and rotation, it tends to overestimate the box scale, whereas our method successfully recovers accurate object dimensions through its iterative scale refinement process. We note that the baseline numbers in Table~\ref{t:HouseCat6D} are taken from the box column of the multi-category HouseCat6D leaderboard rather than from box-specific evaluations, and that these general-purpose methods deliberately do not exploit a box prior. Part of the reported gain therefore stems from our category-specific formulation rather than from a like-for-like comparison.

\vspace{-0.5cm}
\subsubsection{PACE.}

% \begin{table}[t]
% \centering
% \caption{Evaluation of AnyBox on the PACE benchmark. We report 6D pose performance as average precision at IoU = 0.25 and 0.50, and under three pose-accuracy criteria: rotation $\le 20^\circ$, translation $\le 5$ cm, and their conjunction.}
% \label{t:pace}
% \begin{tabular}{@{}l|cc|c|c|c@{}}
% \toprule
%  \multirow{2}{*}{Model} & \multirow{2}{*}{\quad IoU=.25 \quad } & \multirow{2}{*}{IoU=.50 \quad } & \multicolumn{3}{c}{AP} \\
%  \cmidrule(l){4-6} 
%  &  &  & \quad $0:20^\circ$ & \, 5cm \, & $0:20^\circ$/5cm\\ 
%  \midrule
% NOCS \cite{nocs}  & 0 & 0 & 0 & 52.6 & 0 \\
% SGPA \cite{chen2021sgpa}  & 0 & 0 & 0.1 & 19.8 & 0 \\
% HS-Pose \cite{hs-pose}  & 0.8 & 0 & 0 & \textbf{56.5} & 0 \\
% SAR-Net \cite{sarnet}  & 1.3 & 0 & 0 & 46.7 & 0 \\
% CPPF++ \cite{cppf}  & 27.2 & 0 & 0.7 & 21.2 & 0.4 \\
% \textbf{AnyBox (Ours) \,} & \textbf{63.2} & \textbf{10.4} & \textbf{0.8} & 54.1 & \textbf{0.5} \\ \bottomrule
% \end{tabular}\vspace{-0.3cm}
% \end{table}

\begin{table}[t]
\centering
\caption{Evaluation on the PACE benchmark. We report detection AP, pose AP, and ADD-S AP.}
\label{tab:pace_combined}

\small
\setlength{\tabcolsep}{2.5pt}
\renewcommand{\arraystretch}{0.95}

\begin{adjustbox}{max width=\columnwidth}
\begin{tabular}{lccccc|cc}
\toprule
& \multicolumn{2}{c}{\textbf{Det. AP(IoU)}}
& \multicolumn{3}{c|}{\textbf{Pose AP}}
& \multicolumn{2}{c}{\textbf{ADD-S AP (\%)}} \\
\cmidrule(lr){2-3}
\cmidrule(lr){4-6}
\cmidrule(l){7-8}

\textbf{Model}
& 0.25
& 0.50
& $20^\circ$
& 5 cm
& $20^\circ$/5 cm
& 0.1$d$
& 0.15$d$ \\
\midrule

NOCS \cite{nocs}
& 0
& 0
& 0
& 52.6
& 0
& --
& -- \\

SGPA \cite{chen2021sgpa}
& 0
& 0
& 0.1
& 19.8
& 0
& --
& -- \\

HS-Pose \cite{hs-pose}
& 0.8
& 0
& 0
& \textbf{56.5}
& 0
& 5.0
& 31.9 \\

SAR-Net \cite{sarnet}
& 1.3
& 0
& 0
& 46.7
& 0
& 3.6
& 38.8 \\

CPPF++ \cite{cppf}
& 27.2
& 0
& 0.7
& 21.2
& 0.4
& 4.2
& 59.6 \\

\textbf{AnyBox (Ours)}
& \textbf{63.2}
& \textbf{10.4}
& \textbf{0.8}
& 54.1
& \textbf{0.5}
& \textbf{49.2}
& \textbf{67.0} \\

\bottomrule
\end{tabular}
\end{adjustbox}

\vspace{-2mm}
\end{table}

% On the PACE benchmark, in addition to IoU metrics, we report  Average Precision (AP) under three pose accuracy criteria consisting of rotation $\leq 20^\circ$, translation $\leq 5$ cm, and the joint rotation and translation requirement. As shown in Table~\ref{t:pace}, AnyBox achieves the best IoU performance, reaching 63.2 at IoU = 0.25 and 10.4 at IoU = 0.50. Our method has also the highest AP for the rotation $\leq 20^\circ$ criterion (0.8) and for the combined rotation $\leq 20^\circ$ and translation $\leq 5$ cm criterion (0.5), outperforming the previous leading method CPPF++ \cite{cppf}, which reports 27.2, 0.0, 0.7, and 0.4, respectively. The only exception is AP under the 5cm translation criterion, where HS-Pose \cite{hs-pose} ranks first with 56.5 while AnyBox achieves a close second at 54.1. 
On the PACE benchmark, in addition to IoU metrics, we report Average Precision (AP) under three pose accuracy criteria consisting of rotation $\leq 20^\circ$, translation $\leq 5$ cm, and the joint rotation and translation requirement, as well as ADD-S AP at 0.1$d$ and 0.15$d$. As shown in Table~\ref{tab:pace_combined}, AnyBox achieves the best IoU performance, reaching 63.2 at IoU = 0.25 and 10.4 at IoU = 0.50. This corresponds to a 36-point absolute gain over the previous best method, CPPF++ \cite{cppf}, at IoU = 0.25, more than doubling its detection AP, while at the stricter IoU = 0.50 threshold AnyBox is the only method to score above zero. Our method also achieves the highest AP for the rotation $\leq 20^\circ$ criterion (0.8) and the combined rotation $\leq 20^\circ$ and translation $\leq 5$ cm criterion (0.5). The only exception is AP under the 5cm translation criterion, where HS-Pose \cite{hs-pose} ranks first with 56.5 while AnyBox achieves a close second at 54.1. In addition, AnyBox achieves the best ADD-S AP at both thresholds, obtaining 49.2 at 0.1$d$ and 67.0 at 0.15$d$, substantially outperforming previous methods.

\begin{figure}[t]
\centering
\includegraphics[width=1\linewidth]{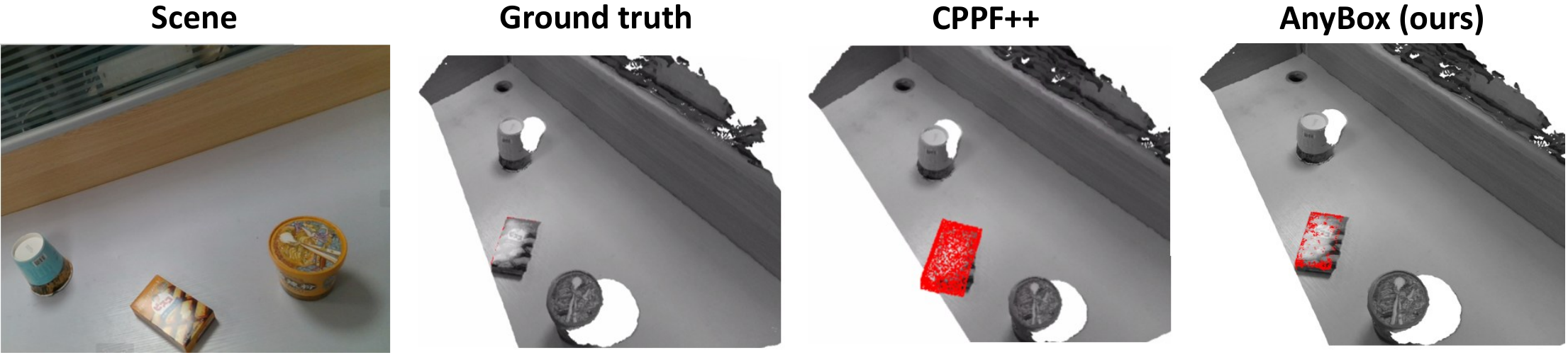}
\caption{Qualitative evaluation of our model, AnyBox on the PACE dataset \cite{pace}.}
\label{fig:example_pace} \vspace{-0.5cm}
\end{figure}

As shown in Fig.~\ref{fig:example_pace}, although CPPF++ is often able to estimate rotation and scale accurately, it frequently fails to recover the correct translation, causing the predicted box to appear unrealistically hovering in front of the scene. In contrast, our method uses a depth consistency filtering step to reject such geometrically inconsistent hypotheses and successfully recovers the correct translation together with accurate rotation and scale.

% \subsubsection{Warehouse6D}
% For this evaluation we consider two SAM6D-based baselines: (i) SAM6D using a canonical category-level template and (ii) SAM6D using the ground-truth CAD model. As shown in Table~\ref{t:baseline}, the category-level template baseline performs poorly under strict overlapping thresholds, achieving precisions of 0.06, 0.02, and 0 at IoU 0.5. 0.7, and 0.9, respectively. In contrast, AnyBox achieves perfect precision at IoU = 0.5 and 0.7, and reaches 0.92 at IoU = 0.9, closely matching SAM6D with ground-truth CAD. The small gap highlights the effectiveness of our approach in recovering accurate instance dimensions while operating without access to instance-specific meshes. More broadly, these results demonstrate that AnyBox effectively bridges the gap between instance-level and category-level pose estimation, combining the flexibility of category-level methods with the geometric accuracy typically associated with instance-level pipelines. Qualitative results are given in Fig. \ref{fig:example_warehouse}.

\subsubsection{Warehouse6D.}
\begin{figure}[t]
  \centering
  \includegraphics[width=1\linewidth]{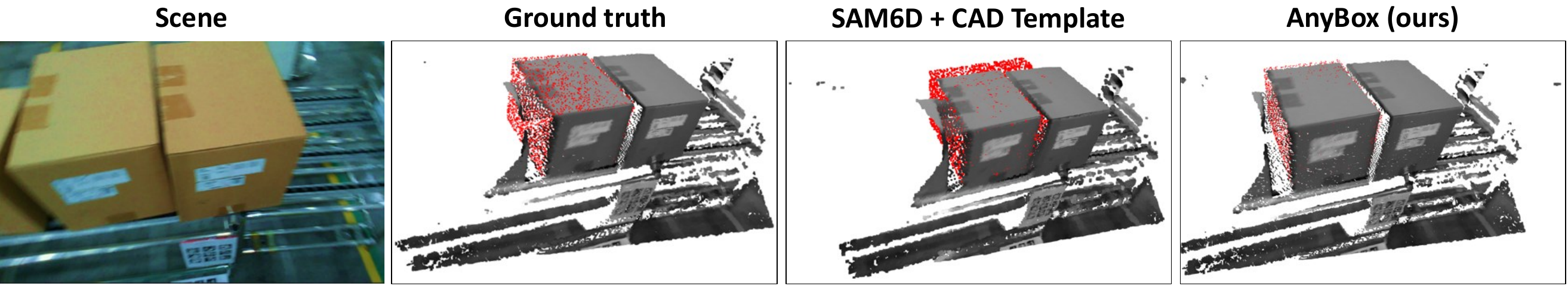}
\caption{Qualitative comparison of AnyBox with the baseline on our warehouse dataset. }
% Estimated poses are shown as red point clouds. \textbf{(a):} a scene from our proprietary warehouse dataset. \textbf{(b):} ground-truth pose. \textbf{(c):} AnyBox estimate. \textbf{(d):} SAM6D estimate using the category CAD template.}
\label{fig:example_warehouse}
\vspace{-0.3cm}
\end{figure}

For this evaluation we consider two SAM6D-based baselines: (i) SAM6D using a canonical category-level template and (ii) SAM6D using the ground-truth CAD model. As shown in Table~\ref{tab:warehouse_combined}, the category-level template baseline performs poorly under strict overlapping thresholds, achieving precisions of 0.06, 0.02, and 0 at IoU 0.5, 0.7, and 0.9, respectively, while also obtaining near-zero pose and ADD-S AP. In contrast, AnyBox achieves perfect precision at IoU = 0.5 and 0.7, and reaches 0.92 at IoU = 0.9, closely matching SAM6D with ground-truth CAD. It further achieves the best performance across all pose and ADD-S metrics. The small gap highlights the effectiveness of our approach in recovering accurate instance dimensions without access to instance-specific meshes, and shows that AnyBox bridges the gap between instance-level and category-level pose estimation, combining the flexibility of the latter with the geometric accuracy of the former. Qualitative results are given in Fig.~\ref{fig:example_warehouse}.

% \begin{table}[tpb]
% \centering
% \caption{Comparison of AnyBox and baselines on the proprietary warehouse dataset.}
% \label{t:baseline} \vspace{-0.2cm}
% \begin{tabular}{@{}l|ccc@{}}
% \toprule
%                          Model  & IoU=.50 \qquad & IoU=.70 \qquad & IoU=.90 \qquad \\ \midrule
% SAM6D \cite{lin2024sam6d} + Canonical CAD \quad & 0.06   & 0.02   & 0.00      \\
% SAM6D \cite{lin2024sam6d} + GT CAD       & 1.00      & 1.00      & \textbf{0.95}   \\
% \textbf{AnyBox (Ours)}                      & \textbf{1.00  }    & \textbf{1.00}      & 0.92   \\ \bottomrule
% \end{tabular}\vspace{-0.1cm}
% \end{table}

\begin{table}[t]
\centering
\caption{Evaluation on the proprietary Warehouse6D benchmark. We report detection AP, pose AP, and ADD-S AP.}
\label{tab:warehouse_combined}

\small
\setlength{\tabcolsep}{2.5pt}
\renewcommand{\arraystretch}{0.95}

\resizebox{\columnwidth}{!}{%
\begin{tabular}{lccc|ccc|cc}
\toprule
& \multicolumn{3}{c|}{\textbf{Det. AP(IoU)}}
& \multicolumn{3}{c|}{\textbf{Pose AP}}
& \multicolumn{2}{c}{\textbf{ADD-S AP (\%)}} \\
\cmidrule(lr){2-4}
\cmidrule(lr){5-7}
\cmidrule(l){8-9}

\textbf{Model}
& 0.50
& 0.70
& 0.90
& $20^\circ$
& 5 cm
& $20^\circ$/5 cm
& 0.1$d$
& 0.15$d$ \\
\midrule

SAM6D + Canonical CAD \cite{lin2024sam6d}
& 0.06
& 0.02
& 0.00
& 0.2
& 0.15
& 0
& 0
& 0 \\

SAM6D + GT CAD \cite{lin2024sam6d}
& 1.00
& 1.00
& \textbf{0.95}
& --
& --
& --
& --
& -- \\

\textbf{AnyBox (Ours)}
& \textbf{1.00}
& \textbf{1.00}
& 0.92
& \textbf{5.4}
& \textbf{7.8}
& \textbf{2.0}
& \textbf{15.0}
& \textbf{82.8} \\

\bottomrule
\end{tabular}
}

\vspace{-2mm}
\end{table}

\vspace{-0.3cm}
% \subsubsection{Ablation studies}

\begin{table}[htbp]
    \centering\vspace{-0.1cm}
    \begin{minipage}{0.48\textwidth}
        \centering
        \captionof{table}{Effect of the depth-consistency filter on 6D pose accuracy on Warehouse6D.}
        \label{t:abl_depth}
        \begin{tabular}{lc}
        \toprule
             Method &  IoU=.80\\
        \midrule
             Without Filter & 0.53 \\
             Depth Consistency Filter & \textbf{0.86} \\
             \bottomrule
        \end{tabular}\vspace{-0.5cm}
    \end{minipage}
    \hfill
    \begin{minipage}{0.48\textwidth}
        \centering
        \captionof{table}{Effect of early stopping on runtime and accuracy on Warehouse6D. Inference time averaged over 50 runs.}
        \label{t:abl_run_time}
        \begin{tabular}{l|cc}
        \toprule
        Method & IoU=.90 & Time(s) \\
        \midrule
        Binary search & \textbf{0.94} & 4.93 ± 2.39 \\
        + Early Stopping & 0.92 &\textbf{1.16 ± 0.74} \\
\bottomrule
\end{tabular}\vspace{-0.5cm}
    \end{minipage}
\end{table}

\subsubsection{Depth-Consistency Filter Ablation.}
We ablate the depth-consistency filter on Warehouse6D by estimating 6D pose with and without this step and measure precision at IoU = 0.80. As shown in Table \ref{t:abl_depth}, adding the filter improves performance from 0.53 to 0.86, an absolute gain of 0.33 and a relative improvement of about 62.3\%. This confirms that rejecting hypotheses whose reprojected mesh is inconsistent with the observed depth substantially improves final pose accuracy.

% \begin{table}[htbp]
%     \centering\vspace{-0.3cm}
%     \caption{Effect of the depth-consistency filter on 6D pose accuracy on Warehouse6D. We report precision at IoU = 0.80.}
%     \label{t:abl_depth}
%     \begin{tabular}{lc}
%     \toprule
%          Method &  IoU=.80\\
%     \midrule
%          Without Filter & 0.53 \\
%          Depth Consistency Filter & \textbf{0.86} \\
%          \bottomrule
%     \end{tabular}\vspace{-0.5cm}
% \end{table}
\subsubsection{Early Stopping Ablation.}
We ablate the effect of adding early stopping at inference time to our pipeline on the warehouse dataset in Table \ref{t:abl_run_time}. We measure mean wall-clock time per frame over 50 runs with and without the early-stopping stage and report 6D pose precision at IoU = 0.90. Early stopping reduces the average runtime from 4.93 ± 2.39 s to 1.16 ± 0.74 s, a 76\% decrease (roughly 4.3× speedup), with a minimal precision drop of 0.02. This shows that early stopping cuts computation substantially, improves runtime stability by lowering variance, and preserves nearly all of the pose accuracy.

% \begin{table}[t]
% \centering
% \caption{Effect of early stopping on runtime and accuracy on the warehouse dataset. We report average precision at IoU = 0.90 and inference time averaged over 50 runs.}
% \label{t:abl_run_time}
% \begin{tabular}{l|cc}
% \toprule
% Method & IoU=.9 & Time(s) \\
% \midrule
% Binary search & \textbf{0.94} & 4.93 ± 2.39 \\
% + Early Stopping & 0.92 &\textbf{1.16 ± 0.74} \\
% \bottomrule
% \end{tabular}\vspace{-0.5cm}
% \end{table}

\begin{figure}[t]
    \centering
    \includegraphics[width=1\linewidth]{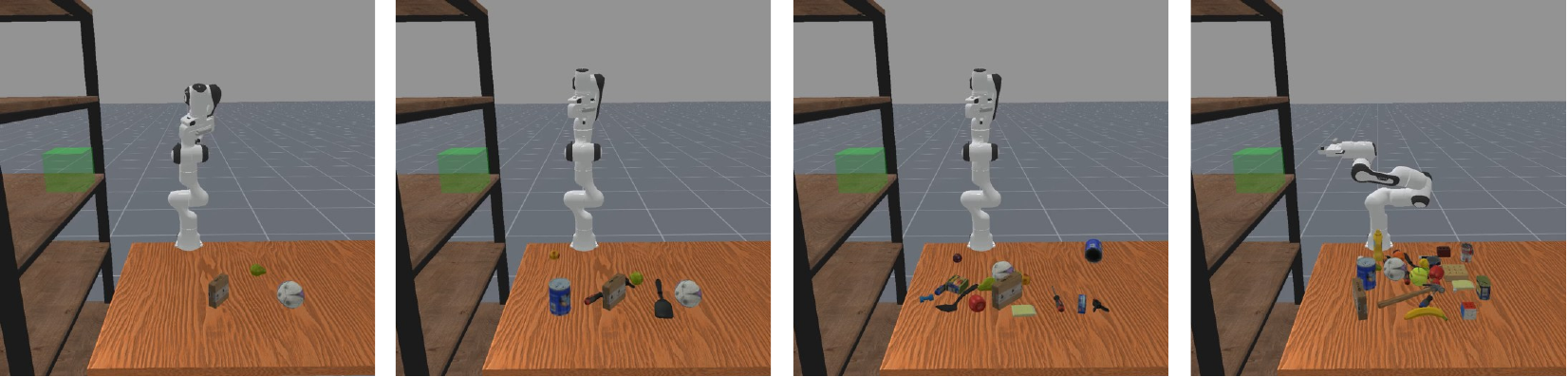}
    \caption{Simulation environment configurations with different numbers of distractors used in the robot experiments.} \vspace{-0.5cm}
    \label{fig:maniskill_env}
\end{figure}

\subsection{Robotic Manipulation in Clutter}
In this section, our goal is to highlight how the gained accuracy in 6D pose estimation using AnyBox translates to improvement in downstream tasks. For this purpose, we use a robotic manipulation task for picking a box from the tabletop and placing it on the shelf. To measure the robustness of our approach, we increasingly populate the environment with non-target objects to increase scene clutter and, consequently, the likelihood of visual distraction and occlusion.

We evaluate AnyBox in a custom simulation environment built using ManiSkill~\cite{maniskill}. Each scene consists of a table containing a target box and a varying number of distractors, along with a predefined goal location represented by a green transparent cube on a shelf adjacent to the table (see Fig. \ref{fig:maniskill_env}). 
The box dimensions are randomized independently along all three axes within $\pm 10$ cm, and the box's roll angle is perturbed within $\pm 45^\circ$. Distractors are randomly selected from the YCB\cite{ycb} object collection, with the number of objects ranging from 1 to 32. The positions of both the target box and distractors are uniformly sampled within a $30\,\text{cm} \times 30\,\text{cm}$ grid on the table (see Fig. \ref{fig:maniskill_env}). We then estimate the pose and drive manipulation with two complementary control paradigms: RRT and VLA.

\subsubsection{RRT Planner.} We first employ a classical Rapidly-exploring Random Tree (RRT)~\cite{RRT} planner, which takes the estimated box pose as its goal and generates a collision-free trajectory to grasp the box from the table and place it at the designated goal location. Because the planner executes the same search-and-track procedure regardless of where the pose comes from, any error in the estimated pose propagates directly into the planned grasp and placement, making this setting a direct, unmediated test of pose accuracy.

\subsubsection{VLA Policy.} We further evaluate $\pi_0$~\cite{pi0}, a state-of-the-art vision-language-action (VLA) model, as a learned, pose-conditioned manipulation policy. In addition to the hand-camera RGB image, the base RGB image, and the robot state, the policy is conditioned on the estimated 6D pose of the target box. Conditioning on the pose in this manner isolates the effect of geometric accuracy on task success, allowing us to assess downstream manipulation success purely as a function of pose estimation quality under progressively increasing clutter, just as with the RRT planner. 
% Details of the $\pi_0$ training procedure and pose-conditioning protocol are provided in the supplementary material.

Table~\ref{tab:manip_results} reports placement success rates under increasing clutter for both control paradigms, comparing AnyBox against the AG-Pose~\cite{lin2024instance} baseline and a ground-truth (GT) oracle that uses the exact object pose retrieved from the simulator. We limit the comparison to AG-Pose because it is the strongest category-level baseline in Table~\ref{t:HouseCat6D} and can be applied to our simulated scenes without dataset-specific retraining. Across both the RRT planner and the $\pi_0$ policy, and at every clutter level, AnyBox consistently outperforms AG-Pose. With the RRT planner, the two methods are comparable under low clutter (a modest 2\% gain at one distractor), but the gap widens substantially as clutter grows, reaching 28\% at four distractors; crucially, AnyBox avoids the sharp degradation exhibited by AG-Pose as distractors accumulate.

\begin{figure}[t]
    \vspace{-0.24cm}
    \centering
    \includegraphics[width=0.8\linewidth]{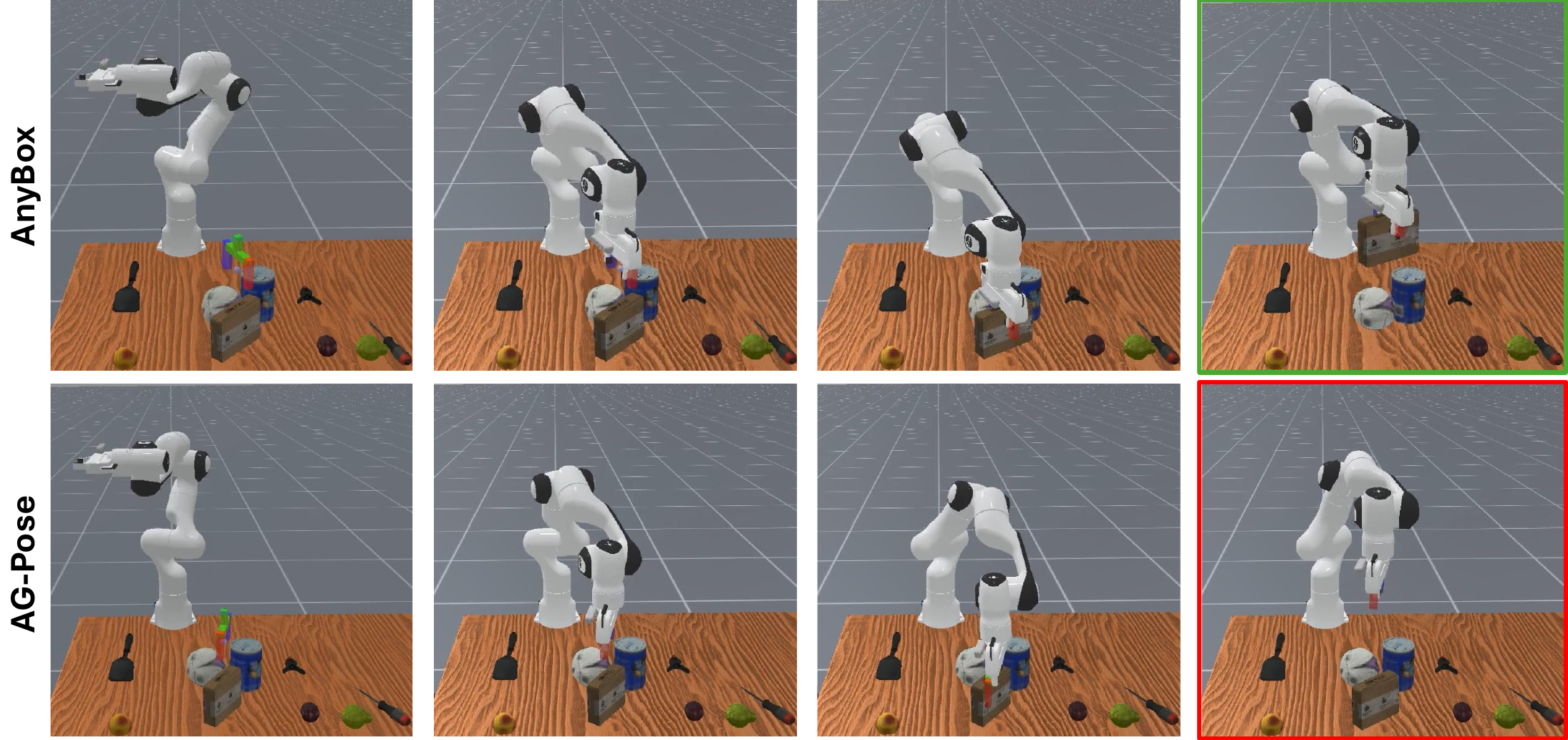}
    \caption{Qualitative examples of robotic manipulation using different pose estimation methods. While the robot successfully grasps the box using the AnyBox-generated pose, it fails to pick the object when using AG-Pose, due to incorrect orientation estimation.} \vspace{-0.6cm}
    \label{fig:robot_qual}
\end{figure}

The learned $\pi_0$ policy follows the same trend, with AnyBox improving over AG-Pose by up to 9\% at two distractors and 6\% at four, while GT conditioning provides an upper bound throughout. We attribute these gains to the refinement mechanisms introduced in our approach, including depth-consistency filtering and explicit scale estimation, which preserve pose accuracy under occlusion and visual interference. The consistency of this improvement across a sampling-based planner and a modern VLA policy indicates that more accurate poses translate into more reliable manipulation regardless of the downstream control paradigm. See Fig.~\ref{fig:robot_qual} for qualitative examples.

\begin{table}[t]
\centering
\caption{Performance of pose estimation methods in a robotic manipulation task under different levels of clutter, using a sampling-based RRT planner and a learned $\pi_0$ VLA policy. Results are reported as success rate.}
\label{tab:manip_results}\vspace{-0.2cm}
\setlength{\tabcolsep}{4pt}
\small
\begin{tabular}{ll cccccc}
& & \multicolumn{6}{c}{\# Distractors} \\
\cmidrule(lr){3-8}
Model & 6D Pose & 1 & 2 & 4 & 8 & 16 & 32 \\
\midrule
\multirow{3}{*}{RRT \cite{RRT}}
& AG-Pose \cite{lin2024instance} & 0.62 & 0.45 & 0.40 & 0.42 & 0.08 & 0.03 \\
& AnyBox & \textbf{0.64} & \textbf{0.59} & \textbf{0.68} & \textbf{0.56} & \textbf{0.09} & \textbf{0.12} \\
\cmidrule(lr){2-8}
& GT & 0.77 & 0.76 & 0.72 & 0.63 & 0.55 & 0.41 \\
\midrule
\multirow{3}{*}{$\pi_0$ \cite{pi0}}
& AG-Pose \cite{lin2024instance} & 0.56 & 0.62 & 0.43 & 0.26 & 0.09 & 0.07 \\
& AnyBox & \textbf{0.64} & \textbf{0.71} & \textbf{0.49} & \textbf{0.30} & \textbf{0.10} & \textbf{0.08} \\
\cmidrule(lr){2-8}
& GT & 0.78 & 0.96 & 0.62 & 0.32 & 0.11 & 0.19 \\
\bottomrule
\end{tabular}\vspace{-0.11cm}
\end{table}

\section{Conclusion and Limitations}
We proposed \textbf{AnyBox}, an efficient 9D pose estimation method designed for box-shaped objects in industrial settings. Building upon a model-based pipeline, AnyBox generalizes across object instances by estimating the instance-specific scale through an efficient binary search procedure. To minimize symmetry and geometric ambiguity, we proposed a depth-consistency filter that rejects pose hypotheses whose reprojected mesh is inconsistent with the observed depth, together with an early-stopping mechanism that transitions to a proportional one-step scale update once the estimated rotation stabilizes, reducing runtime by more than 75\% while preserving final pose accuracy. Extensive evaluations on existing benchmarks, our in-house data, and a downstream robotic manipulation task showed AnyBox significantly outperforms state-of-the-art approaches, improving detection AP by up to 36 points and more than doubling the previous best, while raising manipulation success by 28\% under clutter. These gains come with limitations: AnyBox relies on the upstream SAM segmentation, is most ambiguous on near-cubic boxes, and assumes a rigid cuboid, so deformed or damaged boxes are only approximated by the rescaled template.

\bibliographystyle{splncs04}
\bibliography{main}

% ---------------------------------------------------------------
% Supplementary material (appended for the arXiv version)
\clearpage
\setcounter{figure}{0}
\setcounter{table}{0}
\setcounter{section}{0}
\renewcommand{\thefigure}{S\arabic{figure}}
\renewcommand{\thetable}{S\arabic{table}}
\renewcommand{\thesection}{S\arabic{section}}

\begin{center}
{\large\bfseries AnyBox: Efficient Zero-Shot 9DoF Pose\\
Estimation of Boxes for Robotic Manipulation}\\[0.6em]
{\large Supplementary Material}
\end{center}
\vspace{1em}

\section{Overview}
This supplementary material provides additional experimental details and visual results for \textbf{AnyBox}.  We first introduce the public benchmarks used in our evaluation and discuss the rationale for selecting them for category-level 6D box pose estimation. We then provide additional qualitative examples on these benchmarks. Next, we provide further details about our proprietary Warehouse6D dataset, including dataset statistics and representative samples, followed by additional qualitative results. Finally, we expand the robotic manipulation study by detailing the simulation environment, the randomization protocol, and the pose-conditioning and training setup for our vision-language-action~(VLA) policy experiments.

\section{Benchmark selection.}
We selected the public benchmarks used in this work based on two criteria. 
First, the dataset must contain a sufficient number of box-shaped or packaging-like objects suitable for evaluating category-level box pose estimation. 
Second, the scenes must present realistic challenges such as clutter, occlusion, and diverse viewpoints that reflect conditions encountered in robotic manipulation and warehouse environments.

Several widely used 6D pose estimation benchmarks exist, including LMO~\cite{brachmann2014learning}, YCB-Video~\cite{xiang2018posecnn}, T-LESS~\cite{hodan2017t}, IC-BIN \cite{doumanoglou2016recovering}, and the BOP benchmark suite~\cite{hodan2018bop}. While these datasets have been instrumental in advancing instance-level pose estimation, they contain relatively few objects resembling textureless box-shaped packaging. For example, LMO and YCB-V primarily consist of household objects with distinctive geometry or rich texture, while T-LESS focuses on industrial parts with strong symmetries but few box-like shapes. Similarly, IC-BIN and other BOP datasets emphasize instance-level object categories rather than packaging-style objects with large intra-category variation in scale and appearance. Consequently, the number of suitable textureless box-like objects available for evaluation across these benchmarks is limited, making them less appropriate for studying the category-level box pose estimation problem addressed in this work.

HouseCat6D \cite{housecat} provides cluttered RGB-D scenes with diverse household objects and realistic sensing artifacts. Importantly, it contains multiple box-like objects exhibiting scale variation, weak texture, and partial occlusion. These characteristics make it a suitable benchmark for evaluating the robustness of category-level box pose estimation under challenging visual conditions.

PACE \cite{pace} contains manipulation-centric scenes with packaging objects that closely resemble those found in logistics and robotic picking environments. Compared to many traditional 6D pose benchmarks, PACE includes objects whose geometry and arrangement better reflect real-world manipulation settings. This makes it particularly appropriate for evaluating the downstream utility of pose estimation in robotic tasks involving packaging objects.

Together, these benchmarks satisfy the two selection criteria described above and provide complementary evaluation settings for studying category-level 6D box pose estimation.

\subsection{Public Benchmark Splits}
We evaluate only on the box-relevant subsets of the public benchmarks to match the scope of the proposed method. Since AnyBox is designed specifically for category-level 6D pose estimation of boxes, the evaluation should reflect this formulation rather than broader category-level object pose estimation.

For HouseCat6D, we use the subset of scenes containing box-like household objects. This subset retains the main challenges of the benchmark, including clutter, occlusion, and sensor noise, while ensuring that the evaluation remains aligned with our target object category.

For PACE, we use the packaging and box subset. This subset is especially relevant because it contains objects whose geometry and manipulation characteristics are close to the deployment setting considered in this paper.

\subsection{Additional Qualitative Results on Public Benchmarks}
In this section, we provide additional qualitative examples on the public benchmarks. These results complement the quantitative comparisons reported in the main paper and illustrate the behavior of AnyBox under clutter, occlusion, weak appearance cues, and scale variation.

\begin{figure*}[htb]
    \centering
    \includegraphics[width=\textwidth]{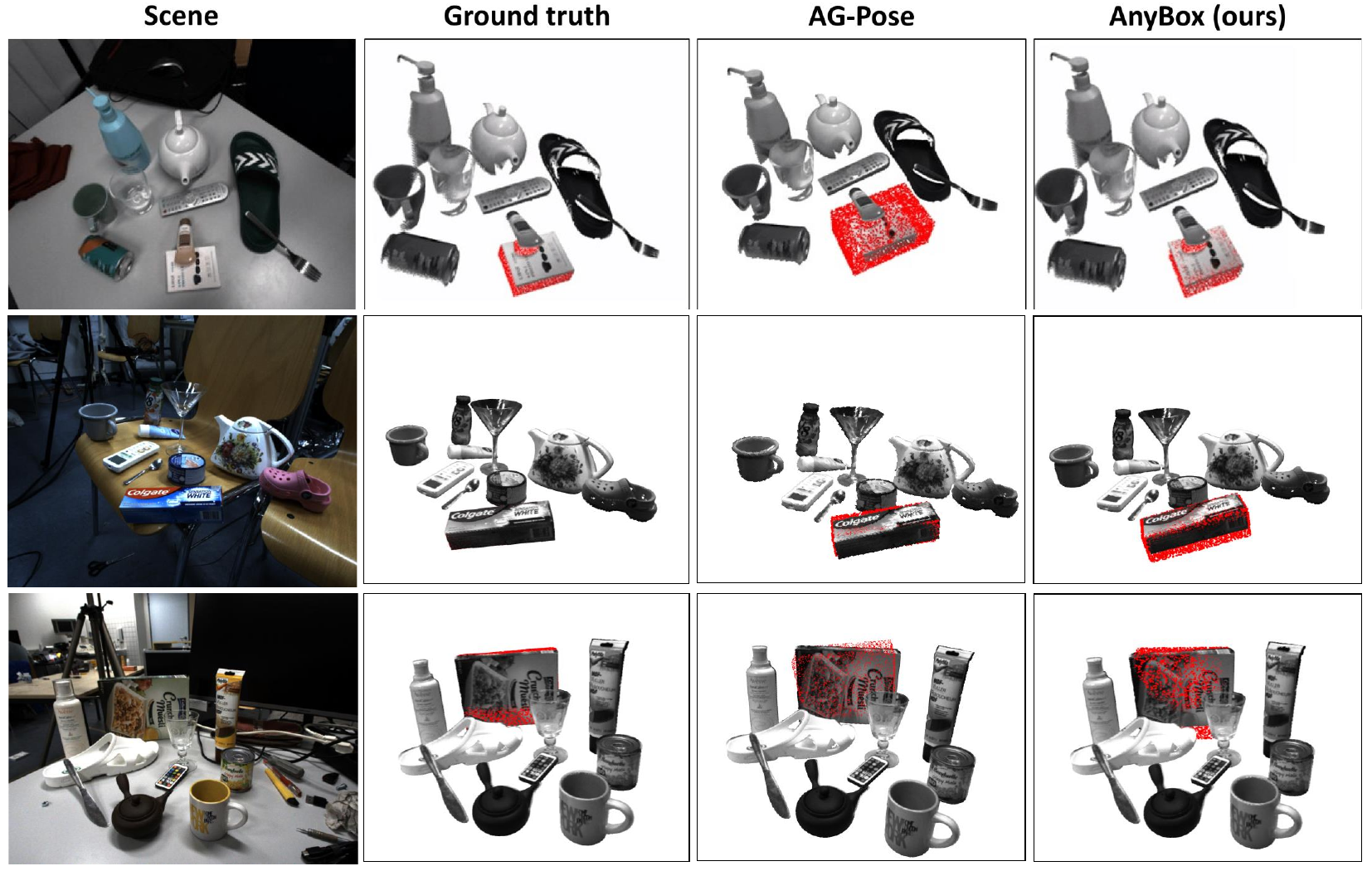}
    \caption{Additional qualitative results on HouseCat6D. The estimated 6D box poses are projected onto the input images. These examples highlight robustness to clutter, occlusion, and instance-level variation in box dimensions.}
    \label{fig:supp_housecat_qual}
\end{figure*}

Figure~\ref{fig:supp_housecat_qual} shows additional qualitative results on HouseCat6D. Across these examples, AnyBox recovers accurate scale, translation, and rotation despite distractors and partial visibility. The examples further illustrate that explicit scale estimation is especially important for box-shaped objects whose dimensions vary substantially across instances.

\begin{figure*}[htb]
    \centering

    \includegraphics[width=\textwidth]{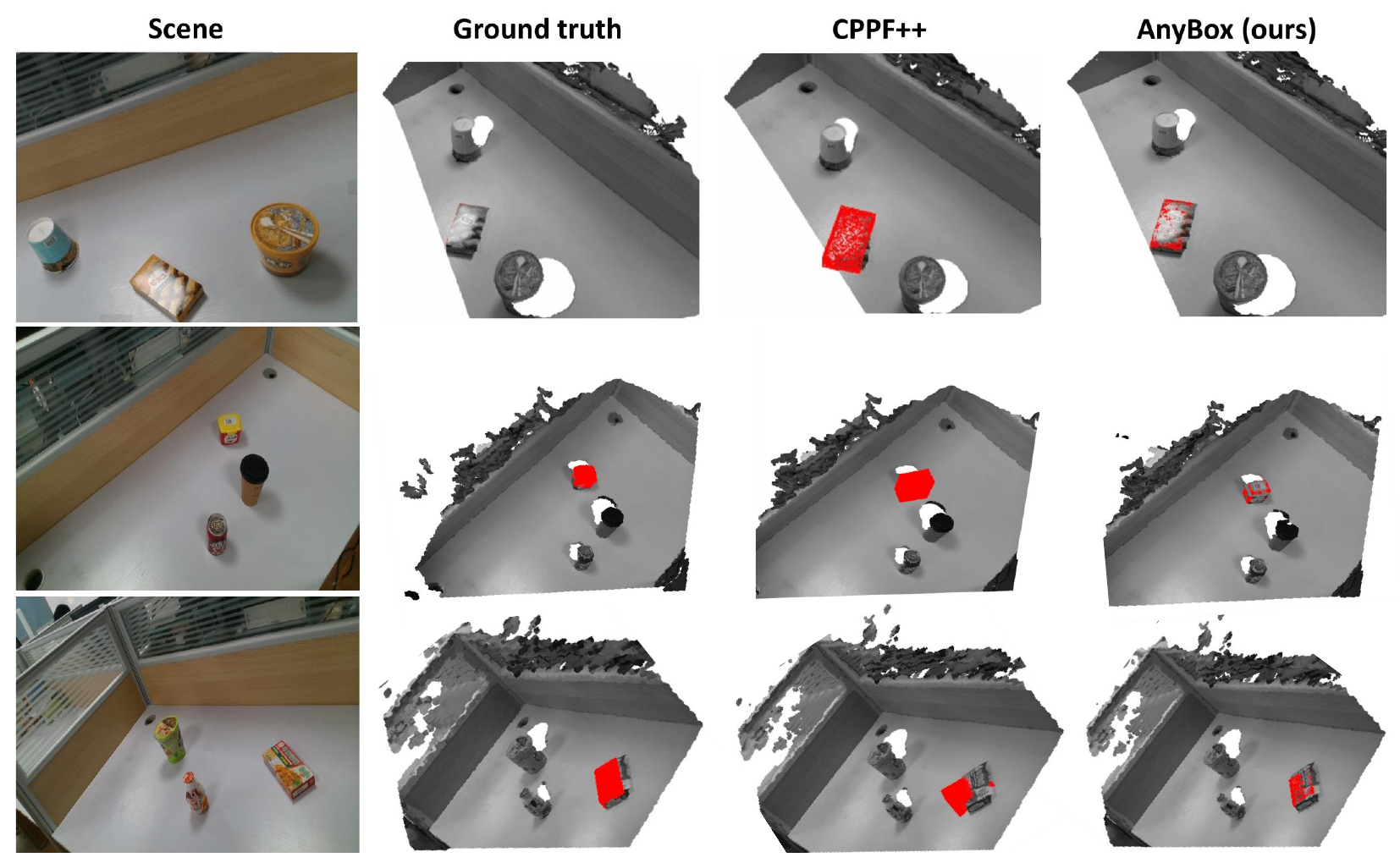}
    \caption{Additional qualitative results on PACE. The examples illustrate that AnyBox remains stable in manipulation-centric scenes with viewpoint ambiguity, clutter, and varying lighting conditions.}
    \label{fig:supp_pace_qual}
\end{figure*}

Figure~\ref{fig:supp_pace_qual} provides additional qualitative examples on PACE. Compared with the examples shown in the main paper, these results further demonstrate the benefit of depth-consistency filtering for removing geometrically implausible hypotheses in ambiguous or partially occluded views.

\section{Warehouse6D Dataset}
\subsection{Dataset Details}
To evaluate the method in a setting that reflects warehouse deployment, we additionally collected Warehouse6D, a proprietary RGB-D dataset focused on box-shaped objects. The dataset contains 2 scenes captured in warehouse and tabletop settings with varying backgrounds, multi-object clutter, and significant scale variation across box instances. In addition to RGB-D images, the dataset provides annotations including 6D object poses, 2D bounding boxes, and segmentation masks for box instances. Compared to the public benchmarks, Warehouse6D better reflects the structure of real logistics scenes, where boxes often appear in piles or stacks, and may be partially occluded by nearby objects.

Table~\ref{tab:supp_warehouse_stats} summarizes the key statistics of the dataset, including the number of scenes, RGB-D frames, annotated object instances, and unique box identities, as well as the train, validation, and test split sizes used in our experiments.

\begin{table}[htb]
\centering
\caption{Summary statistics of the proprietary Warehouse6D dataset.}
\label{tab:supp_warehouse_stats}
\begin{tabular}{l c}
\toprule
Statistic & Value \\
\midrule
Number of scenes & 2 \\
Number of RGB-D frames & 5860 \\
Number of annotated object instances & 17982 \\
Number of unique box identities & 7 \\
Train split & 0.7 \\
Validation split & 0.2 \\
Test split & 0.1 \\
\bottomrule
\end{tabular}
\end{table}

\begin{figure*}[htb]
    \centering
    \includegraphics[width=\textwidth]{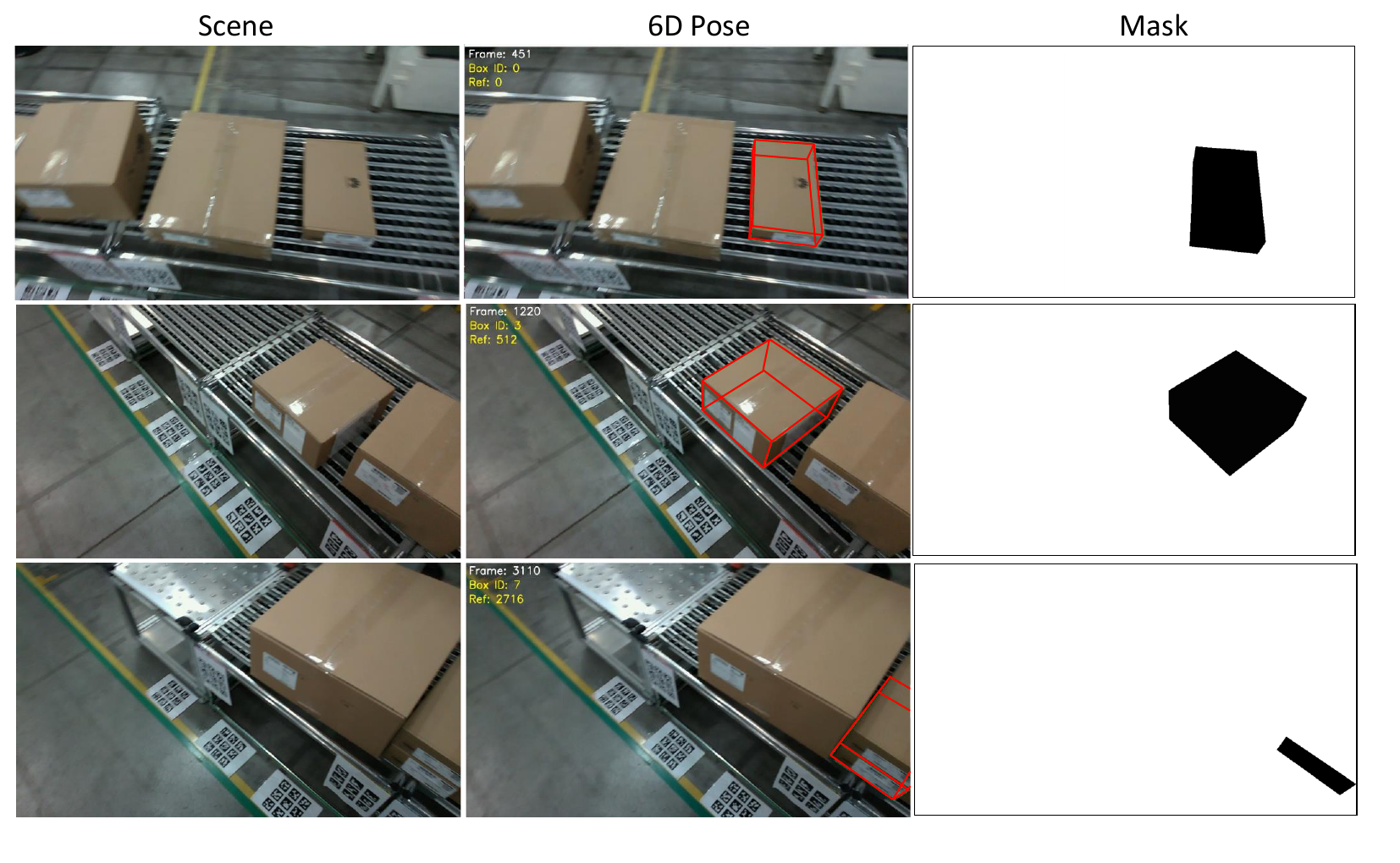}
    \caption{Representative RGB-D samples from Warehouse6D. The dataset includes multiple viewpoints, clutter configurations, and box dimensions that reflect practical warehouse and tabletop settings.}
    \label{fig:supp_warehouse_samples}
\end{figure*}

Figure~\ref{fig:supp_warehouse_samples} shows representative examples from Warehouse6D. These samples illustrate the diversity of viewpoints, object layouts, background conditions, and box dimensions captured in the dataset.

\subsection{Additional Qualitative Results on Warehouse6D}
\begin{figure*}[htb]
    \centering
    \includegraphics[width=\textwidth]{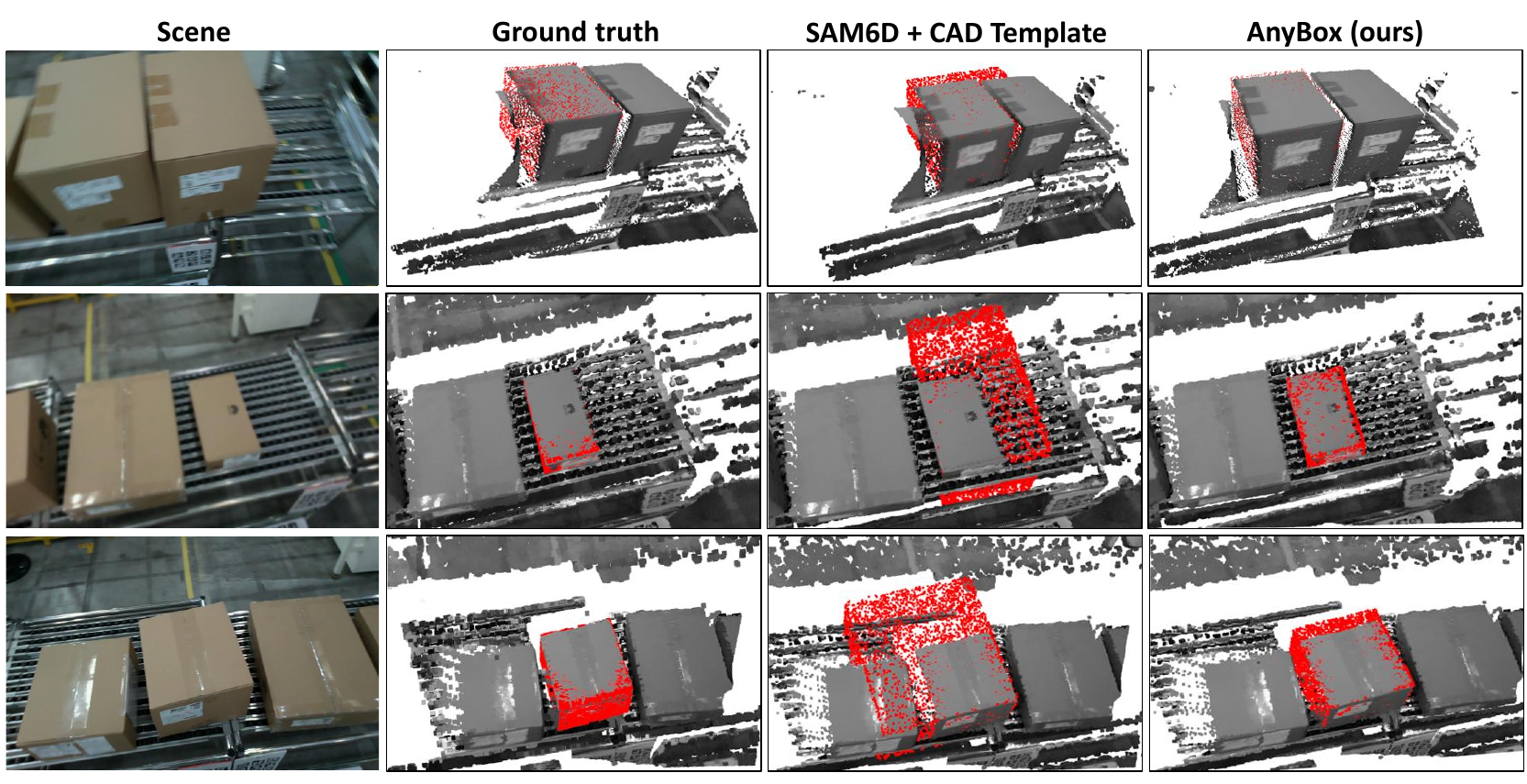}
    \caption{Additional qualitative results on Warehouse6D. The examples show accurate alignment under varying clutter and instance-specific box dimensions.}
    \label{fig:supp_warehouse_qual}
\end{figure*}

Figure~\ref{fig:supp_warehouse_qual} presents additional qualitative results on Warehouse6D. These examples complement the quantitative results in the main paper and further demonstrate that AnyBox recovers accurate 6D poses across varying box sizes and scene layouts, without requiring instance-specific CAD models at inference time.

\section{Robotic Manipulation Experiments}
\subsection{Simulation Environment}
The robotic manipulation experiments are conducted in a custom simulation environment designed to evaluate box picking and shelving under clutter. Each scene contains a target box placed on a table, a shelf with a designated goal location, and a variable number of distractor objects sampled from the YCB object set to introduce geometric and visual diversity around the target.

\subsection{Randomization Protocol}
To evaluate robustness, we randomize the target box and scene configuration along several axes. Specifically, we randomize the box dimensions, roll angle, and placement position on the table. We also vary the number and arrangement of distractor objects across scenes. These randomizations are designed to prevent overfitting to a narrow set of configurations and to stress-test the perception and manipulation pipeline under diverse conditions. Table~\ref{tab:supp_randomization_ranges} summarizes the randomization ranges used in the simulation.

\begin{table}[htb]
\centering
\caption{Randomization ranges used in the robotic simulation. These settings define the variability in the ManiSkill environment to ensure robust policy evaluation.}
\label{tab:supp_randomization_ranges}
\begin{tabular}{l r}
\toprule
Randomized factor & Range / setting \\
\midrule
Box dimensions & $\pm 10$ cm (independent per axis) \\
Box roll angle & $\pm 45^\circ$ (uniform)\\
Box placement on table & $30$ cm $\times 30$ cm grid (uniform) \\
Number of distractors & \{1, 2, 4, 8, 16, 32\} \\
Distractor placement region & $30$ cm $\times 30$ cm grid (uniform) \\
\bottomrule
\end{tabular}
\end{table}

\subsection{Policy Learning with Vision-Language-Action Models}
To evaluate the impact of 6D box pose on downstream manipulation with a learned policy, we utilize $\pi_0$~\cite{pi0}, a state-of-the-art vision-language-action (VLA) model, in a pose-conditioned policy setting. Here, we detail the pose-conditioning protocol and training procedure; the corresponding quantitative results and comparison against the RRT planner are reported in the main paper.

\paragraph{Pose conditioning.} The policy is provided with four inputs: the hand-camera RGB image, the base RGB image, the robot proprioceptive state, and the 6D pose of the target box. The box pose enters the policy as an explicit geometric conditioning signal alongside the visual and proprioceptive streams. Because the visual inputs remain fixed across pose sources, this design isolates how the quality of the geometric information, whether obtained from ground truth or estimation, affects task success. It thereby enables a controlled comparison in which only the source of the conditioning pose (simulator ground truth, AnyBox, or the AG-Pose baseline) varies.

\paragraph{Training details.} To train the policy, we collected 1,000 trajectories in which the robot performs the box-shelving task in a distractor-free environment. We train the model for 75k steps following the recommended training configuration in the $\pi_0$ repository. At inference time, the action horizon is set to 50 to ensure consistent execution.

\subsection{Qualitative Analysis of Failure Cases}

In this section, we provide a qualitative analysis of the failure cases observed when benchmarking RRT-based manipulation across different 6D pose sources. While the quantitative results in the main paper highlight the performance of each method, these visualizations clarify the specific physical and algorithmic challenges encountered during execution.

\begin{figure}[tp]
    \centering
    \includegraphics[width=\columnwidth]{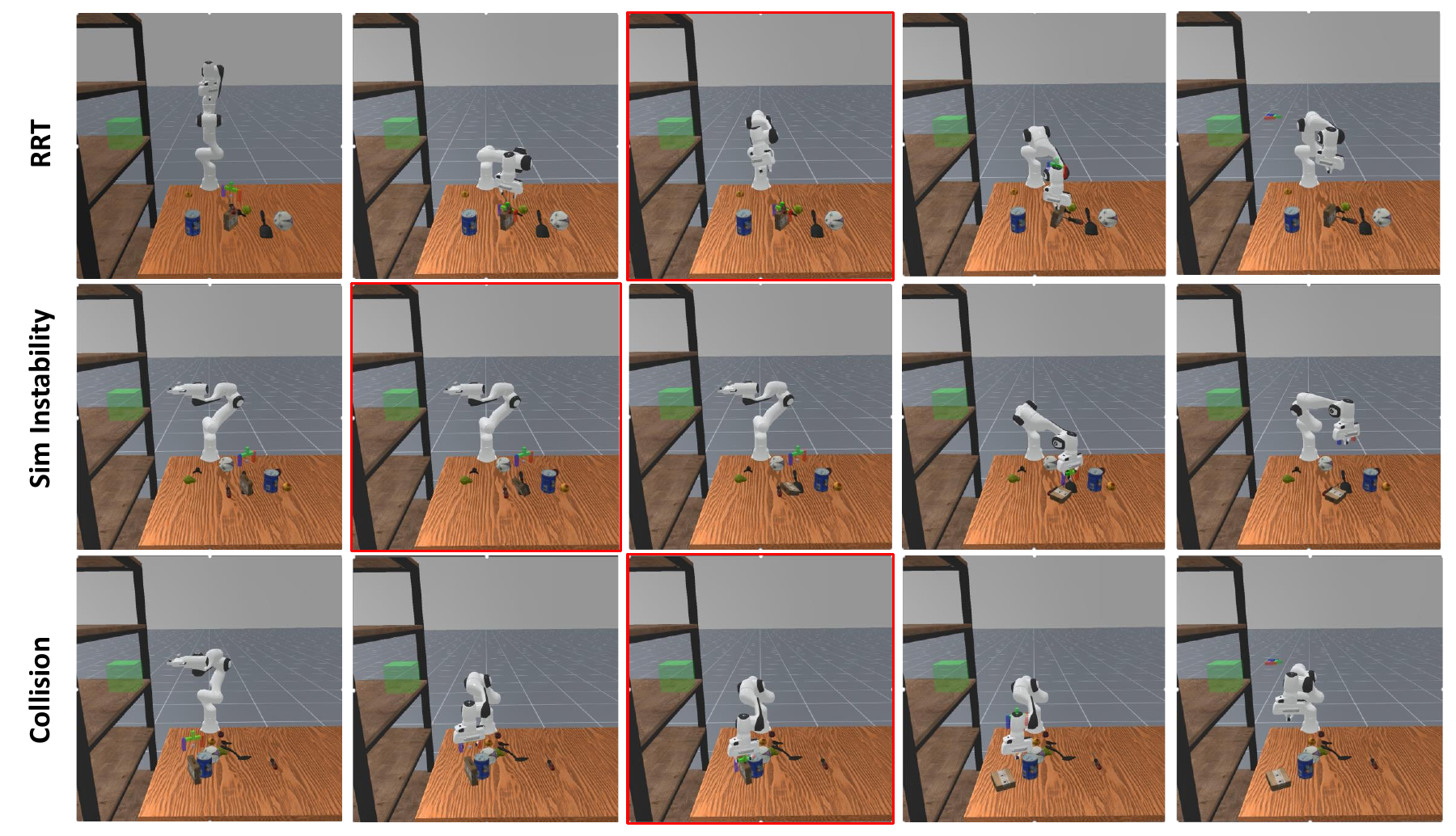}
    \caption{Failure cases using Oracle (GT) poses. Common issues include suboptimal RRT sampling and physical instabilities in the simulator.}
    \label{fig:qual_gt_failure}
\end{figure}

\begin{figure}[tp]
    \centering
    \includegraphics[width=\columnwidth]{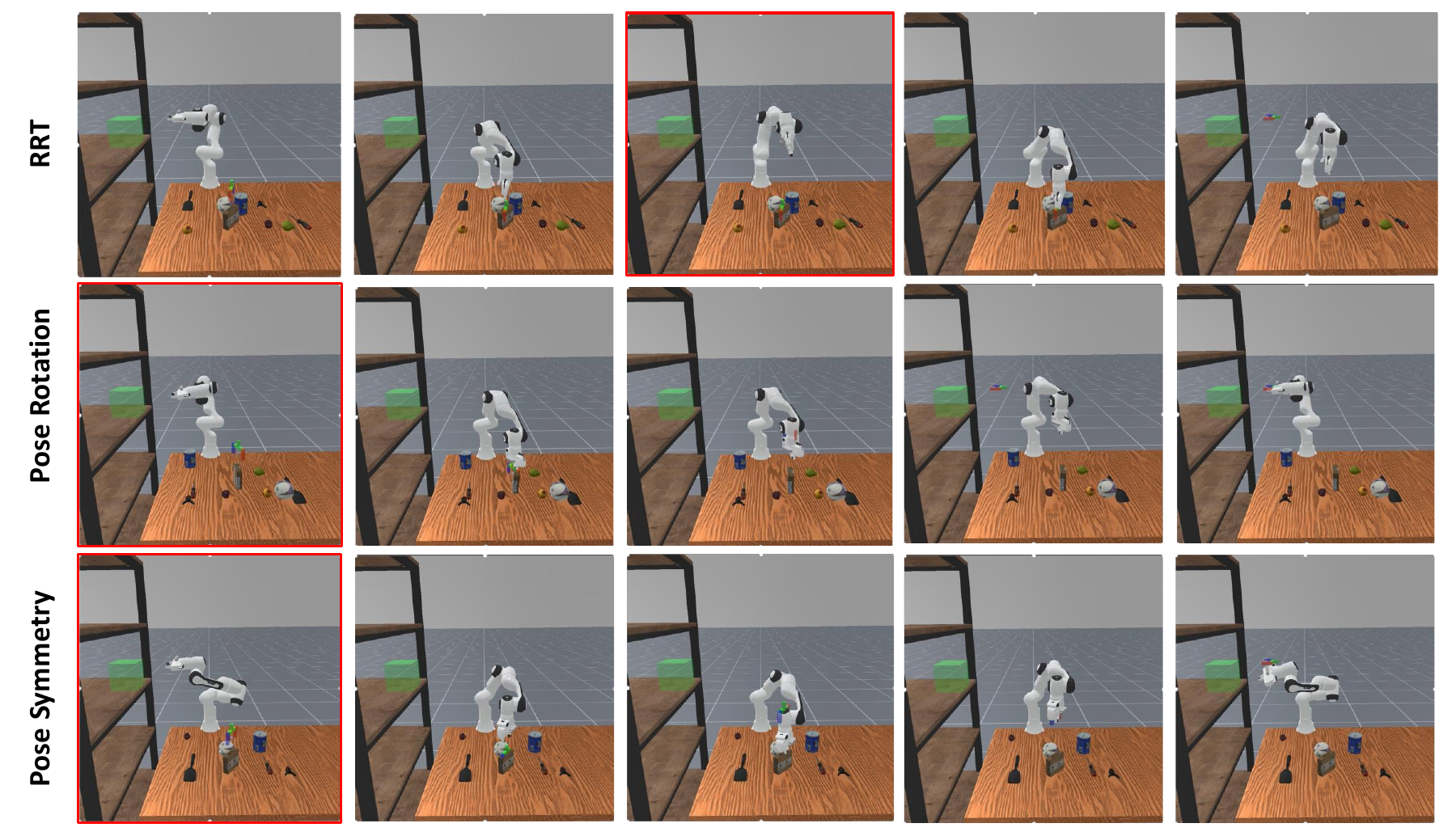}
    \caption{Failure cases using AG-Pose. Errors are primarily driven by 6D pose inaccuracies and symmetry-induced grasp instability.}
    \label{fig:qual_agpose_failure}
\end{figure}

\begin{figure}[tp]
    \centering
    \includegraphics[width=\columnwidth]{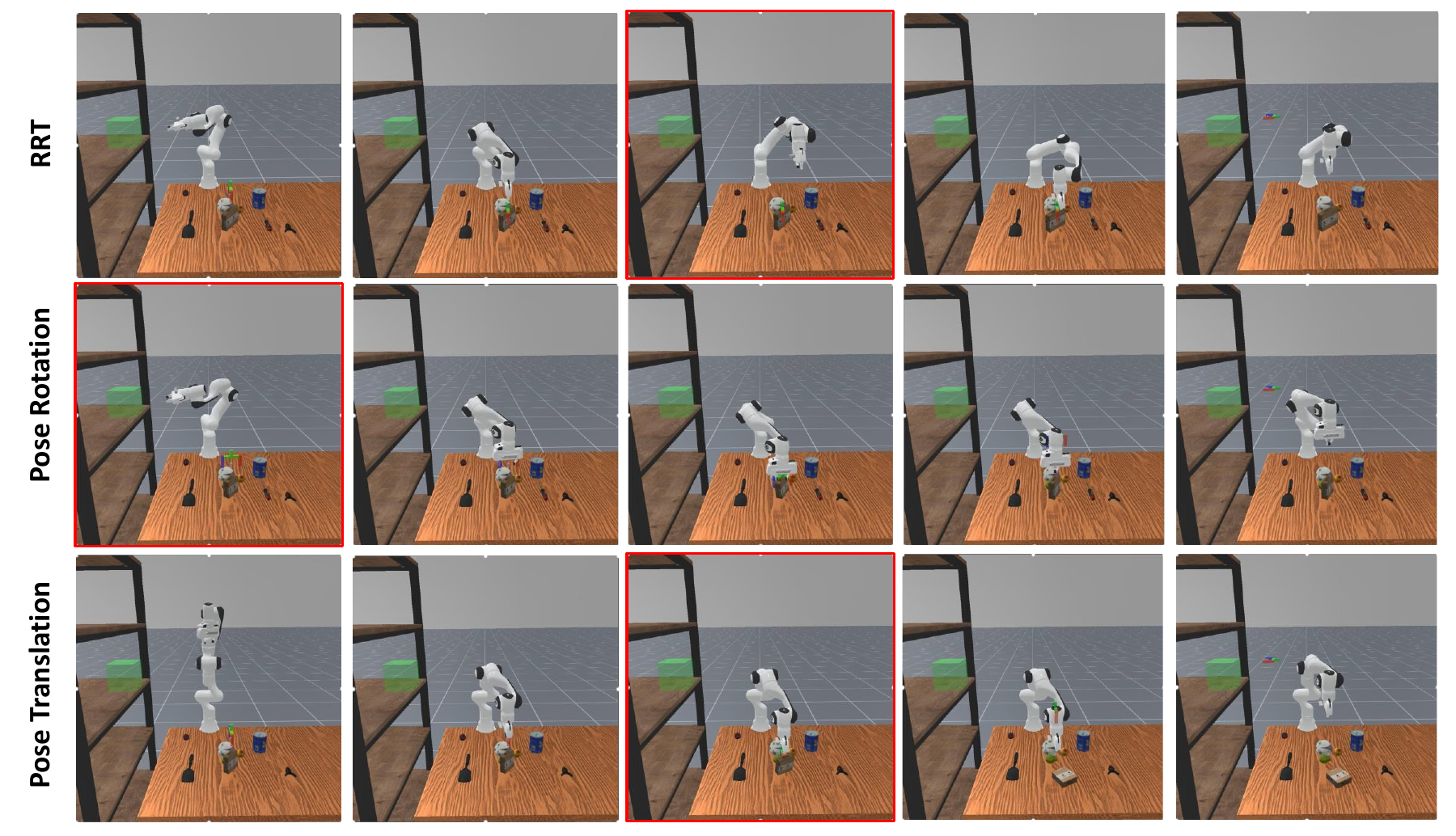}
    \caption{Failure cases using AnyBox. While more robust, failures still occur due to RRT sampling limitations and minor pose estimation residuals.}
    \label{fig:qual_anybox_failure}
\end{figure}

\paragraph{Oracle (GT) Pose Failures}
Even with access to ground-truth 6D poses, the system does not achieve a 100\% success rate. As illustrated in Figure~\ref{fig:qual_gt_failure}, the primary source of error with the oracle pose is the sampling-based nature of the RRT planner, where generated trajectories may occasionally be suboptimal for a successful grasp. Additionally, simulation instability in high-clutter scenes can cause objects to interact unpredictably, positioning the target box in a way that the RRT policy cannot properly resolve. Physical collisions between the robot arm and distractors also frequently prevent a successful grasp despite the accuracy of the pose information.

\paragraph{AG-Pose Failures}
The failure cases for AG-Pose, shown in Figure~\ref{fig:qual_agpose_failure}, highlight the compounding effect of pose estimation errors on sampling-based motion planning. Inaccurate pose estimates frequently lead to manipulation failures in which the gripper misses the target or collides with the box edges. Furthermore, AG-Pose struggles with box and pose symmetry, often producing unstable or physically unreachable grasp poses. These issues are exacerbated by RRT sampling, which cannot compensate for the underlying geometric misalignment.

\paragraph{AnyBox Failures}
While AnyBox demonstrates strong robustness overall, certain failure modes persist, as depicted in Figure~\ref{fig:qual_anybox_failure}. The stochastic nature of RRT sampling remains a bottleneck, as the planner may fail to find a valid path even when provided with a high-quality pose estimate. We also observe that residual errors in the rotation and translation components of the estimated 6D pose can lead to manipulation failures, particularly when slight misalignment causes the box to slip during the lifting or shelving phase.

\section{Conclusion}
This supplementary material provided additional evidence for the claims made in the main paper. We clarified the choice of public benchmarks, presented more details and qualitative results for the proprietary Warehouse6D dataset, and expanded the robotic manipulation study with environment details, randomization analysis, 6D pose visualizations, and the pose-conditioning and training setup for our VLA policy experiments. Together, these results further support the effectiveness of AnyBox for accurate category-level 6D pose estimation of boxes and its value for downstream robotic manipulation in cluttered environments.

\end{document}